\documentclass{article} 
\usepackage{iclr2022_conference,times}

\usepackage[acronym]{glossaries}
\glsdisablehyper

\usepackage{hyperref}
\usepackage{url}

\usepackage[pdftex]{graphicx}
\graphicspath{{figures/}}
\usepackage{amssymb}
\usepackage{subcaption}
\usepackage{caption}
\usepackage{placeins}

\usepackage{pgfplots}
\pgfplotsset{width=10cm,compat=1.9}

\usepackage{amsmath,amsfonts,bm}

\def\eqref#1{equation~\ref{#1}}

\def\1{\bm{1}}

\def\vh{{\bm{h}}}

\def\vr{{\bm{r}}}

\def\mW{{\bm{W}}}

\DeclareMathAlphabet{\mathsfit}{\encodingdefault}{\sfdefault}{m}{sl}
\SetMathAlphabet{\mathsfit}{bold}{\encodingdefault}{\sfdefault}{bx}{n}

\newacronym{sundae}{SUNDAE}{step-unrolled denoising autoencoders}
\newacronym{ar}{AR}{autoregressive}
\newacronym{nar}{NAR}{non-autoregressive}

\newacronym{rnn}{RNN}{recurrent neural network}
\newacronym{lstm}{LSTM}{long short-term memory recurrent neural network}
\newacronym{gru}{GRU}{gated recurrent unit}

\newacronym{gan}{GAN}{generative adversarial network}
\newacronym{vae}{VAE}{variational auto-encoder}
\newacronym{vq}{VQ}{vector-quantization}
\newacronym{sbm}{SBM}{score-based model}
\newacronym{ddpm}{DDPM}{denoising diffusion probabilistic model}
\newacronym{ebm}{EBM}{energy-based model}

\newacronym{vqgan}{VQ-GAN}{vector-quantized \acrshort{gan}}
\newacronym{vqvae}{VQ-VAE}{vector-quantized \acrshort{vae}}

\newacronym{flop}{FLOP}{floating point operation}

\newcommand{\real}[1]{\mathbb{R}^{#1}}

\newcommand{\imageDataset}{\mathcal{X}}
\newcommand{\image}{\mathbf{x}}
\newcommand{\sample}{\mathbf{y}}
\newcommand{\pixel}[1]{x_{#1}}

\newcommand{\latentDataset}{\mathcal{L}}
\newcommand{\latent}{\mathbf{z}}
\newcommand{\latentSpace}{Z}
\newcommand{\latentHeight}{h}
\newcommand{\latentWidth}{w}

\newcommand{\vqganEncoder}{E}
\newcommand{\vqganDecoder}{G}
\newcommand{\vqganCodebook}{\mathcal{C}}
\newcommand{\vqganDownsample}{f}
\newcommand{\vqganNbLatents}{v}
\newcommand{\codebookVector}{\mathbf{e}}

\newcommand{\hourglassRate}{k}
\newcommand{\hourglassResample}{\mW}

\newcommand{\sundaeParameters}{\theta}
\newcommand{\markovSteps}{T}
\newcommand{\sundae}{f_\sundaeParameters}
\newcommand{\corruptionThreshold}{\textit{t}}

\newcommand{\corruptionDistribution}{q}

\newcommand{\lossFunction}[1]{L^{(#1)}(\sundaeParameters)}
\newcommand{\pixelMask}{m_p}
\newcommand{\vqMask}{m_\textrm{vq}}

\newcommand{\temperature}{\tau}
\newcommand{\bigO}[1]{\mathcal{O}(#1)}

\title{Megapixel Image Generation with Step-Unrolled Denoising Autoencoders}

\author{Alex F. McKinney\thanks{Corresponding Author:
    \texttt{alex.f.mckinney@gmail.com}} \;\& Chris G. Willcocks \\
Department of Computer Science,
Durham University,
Durham, UK \\
}

\iclrfinalcopy 
\begin{document}

\maketitle

\begin{abstract}
    An ongoing trend in generative modelling research has been to push sample
resolutions higher whilst simultaneously reducing computational requirements for
training and sampling. We aim to push this trend further via the combination of
techniques---each component representing the current pinnacle of efficiency in
their respective areas. These include \acrfull{vqgan}, a \gls{vq} model capable
of high levels of lossy---but perceptually insignificant---compression;
hourglass transformers, a highly scaleable self-attention model; and
\acrfull{sundae}, a \acrfull{nar} text generative model. Unexpectedly, our
method highlights weaknesses in the original formulation of hourglass
transformers when applied to multidimensional data. In light of this, we propose
modifications to the resampling mechanism, applicable in any task applying
hierarchical transformers to multidimensional data. Additionally, we demonstrate
the scalability of \acrshort{sundae} to long sequence lengths---four times
longer than prior work. Our proposed framework scales to high-resolutions ($1024
\times 1024$) and trains quickly (2-4 days). Crucially, the trained model
produces diverse and realistic megapixel samples in approximately 2 seconds on a
consumer-grade GPU (GTX 1080Ti). In general, the framework is flexible:
supporting an arbitrary number of sampling steps, sample-wise self-stopping,
self-correction capabilities, conditional generation, and a \acrshort{nar}
formulation that allows for arbitrary inpainting masks. We obtain FID scores of
10.56 on FFHQ256---close to the original \acrshort{vqgan} in less than half the
sampling steps---and 21.85 on FFHQ1024 in only $100$ sampling steps.

\end{abstract}

\begin{figure}[ht]
    \centering
    \makebox[\textwidth]{
        \includegraphics[width=1.2\textwidth]{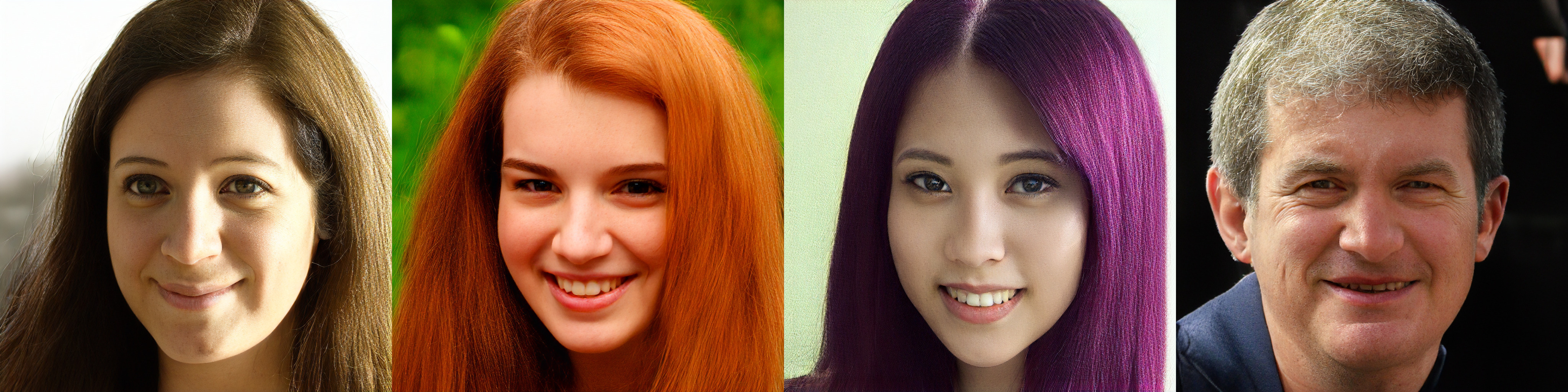}
    }
    \caption{
        Samples produced using our non-autoregressive approach. Each $1024
        \times 1024$ sample was generated in $\approx 2$ seconds on a GTX 1080Ti
       ---including both discrete latent sampling and subsequent VQ-GAN
        decoding. At this resolution, autoregressive and (non-adversarial)
        non-autoregressive models take minutes to sample, or simply do not scale
        to this resolution. 
    }
    \label{fig:main}
\end{figure}

\section{Introduction}
The ideal deep generative model would satisfy three key requirements:
high-quality samples, mode coverage resulting in high sample diversity, and
computationally inexpensive sampling. Arguably, there are other desirable
properties such as a meaningful latent space, exact likelihood calculation, and
controllable generation. Nonetheless, no current generative model satisfies all
three key requirements---let alone additional attractive properties---forming
the so-called generative modelling trilemma~\citep{xiao2021trilemma} that
dominates modern generative modelling research.

For example, \glspl{gan}~\citep{goodfellow2014gan} excel at high-quality and
fast sampling, but are unstable to train and susceptible to mode
collapse~\citep{xiao2021trilemma}. Models such as Image
Transformer~\citep{parmar2018image}, \glspl{ddpm}~\citep{ho2020ddpm}, and
\glspl{sbm}~\citep{song2019sbm} are stable to train, have mode covering
characteristics, and produce high-quality samples. However, they require many
network evaluations to produce a single sample.
\Glspl{vae}~\citep{kingma2013vae} permit single step sampling, but fails to
produce samples of competitive fidelity. Normalizing flows~\cite{dinh2014nice}
offer exact likelihood calculation, but have a restrictive architecture that
makes it parameter inefficient and hard to scale.
\Acrfull{vq}~\cite{oord2017vqvae} image models help alleviate computational
costs, but mandate a two-stage training scheme and one or more additional
models.

This overview of generative modelling demonstrates that no current approach
satisfies all three requirements. This motivates research into explicitly
addressing this trilemma. In this work, we move towards such a solution,
beginning from existing work applying generative models to discrete latents.
This provides an excellent starting point in terms of sample quality and mode
coverage, but with slow sampling speeds despite operating in a small latent
space. We address this by sampling latents using a \gls{nar} generative model to
close the gap, sampling speed wise, with models such as \glspl{gan}.
Specifically, we use \gls{sundae}~\citep{savinov2022stepunrolled} to denoise
samples from a uniform prior into samples from the discrete latent space defined
by a trained \gls{vqgan}. We find that \gls{sundae} is an effective discrete
prior over \gls{vqgan} representations, even on sequence lengths greater than
previously evaluated on.

\Gls{sundae} has only previously be applied to language modelling
tasks~\citep{savinov2022stepunrolled} using
transformers~\citep{vaswani2017attention} to implement their model. Parallel
work introduced a drastically more efficient variant---the hourglass
transformer~\citep{nawrot2021hierarchical}---leveraging a hierarchical
architecture targeting language modelling. Though able to be applied to discrete
latent modelling, we propose a number of improvements that improve performance
on multidimensional data, including modifications to resampling operations and
introduction of axial positional embeddings~\citep{su2021roformer}. Though
evaluated on discrete latents, the modifications are applicable to any
multidimensional data, valuable outside a generative modelling context.
Hence, this work further demonstrates the efficacy of hierarchical transformers
outside of language tasks.

Given a fast sampling and efficient transformer architecture, we now possess a
highly scaleable generative model, with respect to number of layers and spatial
resolution of the latent input. Only a minority of the layers are operating at
the same resolution as the input, reducing the impact of costly self-attention.
Conversely, we cheaply scale the number of layers by adding layers only at the
downsampled resolution, allowing for considerably larger models with minor
additional cost. To demonstrate this scalability, we train a \gls{vqgan}
operating on $1024 \times 1024$ images and apply our framework to the resulting
discrete latents. This results in the \textbf{synthesis of megapixel images in
as few as two seconds} on a consumer-grade GPU. To our knowledge, this is the
largest \gls{vqgan} trained in terms of input size, and the fastest sampling
non-adversarial generative framework at this image resolution. This framework
includes a number of additional desirable features, such as self-correction,
sample-wise self-stopping, conditional generation, and flexible inpainting.

\section{Related Work}
This work builds upon much prior research into powerful deep generative models,
self-supervised representation methods, and efficient transformer architectures,
some of which we discuss in this section. For a full review on generative
modelling we direct the reader to \citet{bondtaylor2021review}, and on
\gls{sundae} and hourglass transformers to \citet{savinov2022stepunrolled} and
\citet{nawrot2021hierarchical} respectively.

\begin{figure}[ht!]
    \centering
    \makebox[\textwidth]{\includegraphics[width=1.0\textwidth]{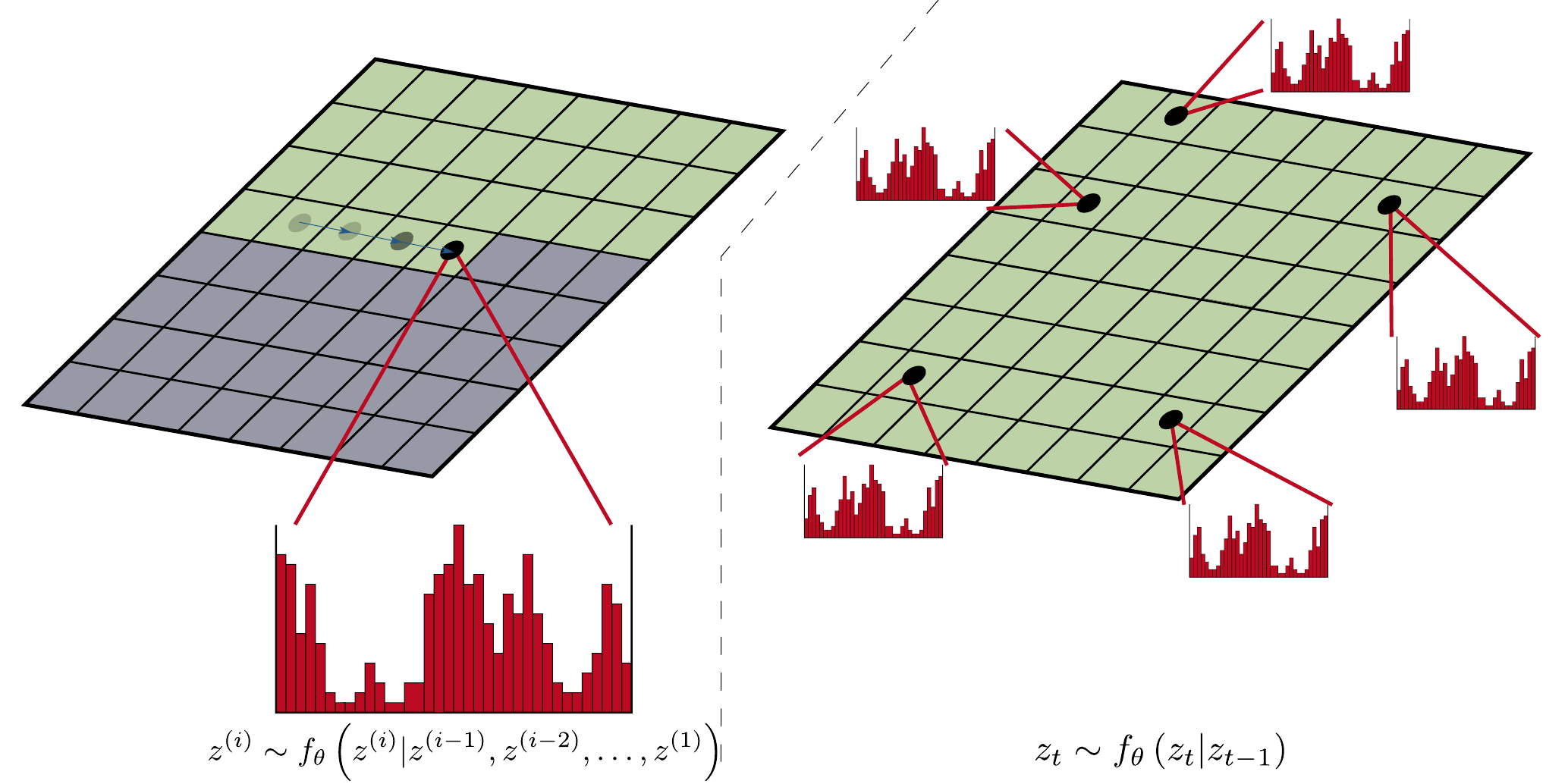}}
    \caption{
        \textbf{Left:} Visualization of \acrfull{ar} sampling. \Gls{ar} sampling
        proceeds one element at a time, meaning the number of sampling steps is
        equal to the dimensionality of the input and can only make use of past
        context (indicated by green positions). 
        \textbf{Right:} Visualization of \acrfull{nar} sampling. \Gls{nar}
        sampling samples an arbitrary number of elements in parallel, allowing
        for full context and potentially faster sampling.
    }
\end{figure}

\subsection{Autoregressive Generative Models}
One major deep generative model family are \acrfull{ar} models, characterised by
a training and inference process based on the probabilistic chain rule. During
training, they learn to maximise the likelihood of the data they are trained on,
which leads to excellent mode coverage. Prior work using these methods resulted
in impressive results in terms of sample quality and diversity, but are
ultimately impractical in most real world applications due to their slow
sampling speed.

The slow speed is due to their sequential nature, defined by the chain
rule of probability. Given an input $\image = \{ \pixel{1}, \pixel{2}, \dots,
\pixel{N} \}$, an \gls{ar} model $p_\theta(\cdot)$ generates new
samples sequentially:
\begin{equation}\label{eq:ar}
    p_\theta(\image) = p_\theta(\pixel{1}, \dots, \pixel{N}) =
    \prod\limits^{N}_{i=1} p_\theta(\pixel{i} \vert \pixel{1}, \dots, \pixel{i-1})
\end{equation}
meaning that the number of sampling steps is equal to the size of the
decomposition of $\image$, giving an iteration complexity of $\bigO{N}$.

For certain tasks, the ordering of the decomposition of $\image$ is obvious, for
example on text or speech. For images this is less clear, but typically a raster
scan ordering is used~\citep{parmar2018image}. Certain \gls{ar} models are
order-agnostic~\citep{hoogeboom2021autoregressive}, allowing for arbitrary
ordering to be used during training and inference.

Early examples of \gls{ar} models include \glspl{rnn} and derivatives
\glspl{gru}~\citep{cho2014gru} and \glspl{lstm}~\citep{hoch1997lstm}, typically
applied to text, audio, and other time series data. Later work introduced models
on images, such as PixelRNN and PixelCNN~\citep{oord2016pixelrnn} and
PixelSnail~\citep{chen2017snail}. The introduction of transformer
architectures~\citep{vaswani2017attention} naturally led to applications in
image generation~\citep{parmar2018image}, extending even to zero-shot,
text-to-image models~\citep{ramesh2021dalle}. In all cases, the iteration
complexity is still $\bigO{N}$, a property intrinsic to \gls{ar} models.

\subsection{Non-autoregressive Generative Models}
\Acrfull{nar} generative models include \glspl{gan}~\citep{goodfellow2014gan},
\glspl{vae}~\citep{kingma2013vae}, \glspl{sbm}~\citep{song2019sbm},
\glspl{ddpm}~\citep{ho2020ddpm}, and flow-based models~\citep{dinh2014nice}. The
number of sampling steps in \gls{nar} models is not directly tied to the data
dimensionality, however the actual number of steps varies greatly: from
single-step generation in \glspl{gan} and \glspl{vae}, to many thousands in early
\glspl{ddpm} and \glspl{sbm}.

The single-step sampling coupled with high-fidelity results makes \glspl{gan}
the de facto standard for generative models in practical applications. However,
\glspl{gan} are plagued with issues such as unstable training and tendency to
collapse onto modes of the target distribution, rather than model the entire
target distribution. This is due to not directly optimising for likelihood, but
rather using an adversarial loss as a proxy objective. This motivates research
into producing other fast-sampling generative models that do not have these
issues. Clearly, \gls{ar} models will never satisfy this requirement due to
their $\bigO{n}$ iteration complexity, leaving improving existing, or creating
new, \gls{nar} models the only avenue available.

\subsection{Step-unrolled Denoising Autoencoder}
\label{subsec:sundae}

\Gls{sundae}~\citep{savinov2022stepunrolled} is a \gls{nar} text generative
model evaluated on three language modelling tasks: unconditional
text-generation, inpainting of Python code, and machine translation---setting a
new state-of-the-art among \gls{nar} models for the latter. It is capable of
fast sampling, producing high quality text samples in as few as 10 steps.

It is trained using a denoising objective, akin to 
BERT's objective~\citep{wang2019bert} but with multiple denoising steps.
Given a uniform prior $p_0$ over some space $\latentSpace = \{1, \dots,
\vqganNbLatents\}^N$ where $N$ is the size of the space and $v$ is the
vocabulary size, consider the Markov process $\latent_t \sim \sundae(\cdot \vert
\latent_{t-1})$ where $\sundae$ is a neural network parameterised by
$\sundaeParameters$, then $\{\latent_t\}_t$ forms a Markov chain. This gives a
$t$-step transition function: 
\begin{equation}\label{eq:markov} p_t(\latent_t
    \vert \latent_0) = \sum\limits_{\latent_1, \dots, \latent_{t-1} \in
    \latentSpace} \prod\limits^t_{s=1} \sundae(\latent_s | \latent_{s-1})
\end{equation}
and, given a constant number of steps
$\markovSteps$, our model distribution $p_\markovSteps(\latent_\markovSteps
\vert \latent_0)p_0(\latent_0)$---which is clearly intractable.

Instead, \gls{sundae} uses an \textit{unrolled denoising} training method that
uses a far lower $\markovSteps$ than is used for sampling. To compensate, they
unroll the Markov chain to start from corrupted data produced by a
\textit{corruption distribution} $\latent_0 \sim \corruptionDistribution(\cdot
\vert \latent)$ rather than from the prior $p_0$ so the model during training
sees inputs alike those seen during the full unroll at sample
time~\citep{savinov2022stepunrolled}. Typically, $\markovSteps = 2$ is used
during training, as a single step would be similar to BERT's
objective~\citep{devlin2019bert} which would not be performant as seen in
earlier work using BERT as a random field language model~\citep{wang2019bert}.

The training objective of \gls{sundae} is the average of all cross-entropy
losses on the chain after $t$ steps, which is shown to form an upper bound on
the actual negative log-likelihood~\citep{savinov2022stepunrolled}. Increasing
$\markovSteps$ leads to a minor improvement in performance, but slows down
training and increases memory usage.

One advantage of this approach is that sampling starts from random tokens,
rather than a ``masking''
token~\citep{bondtaylor2021unleashing,austin2021structured}. Unmasking
approaches means that $\markovSteps \leq N$ as at minimum, one token is unmasked
per step. Additionally, a random prior allows the model to ``change its mind''
about previously predicted elements during sampling, permitting fine adjustments
and correction of errors.

\subsection{Vector Quantized Image Modelling}
\begin{figure}[ht!]
    \centering
    \makebox[\textwidth]{\includegraphics[width=1.1\textwidth]{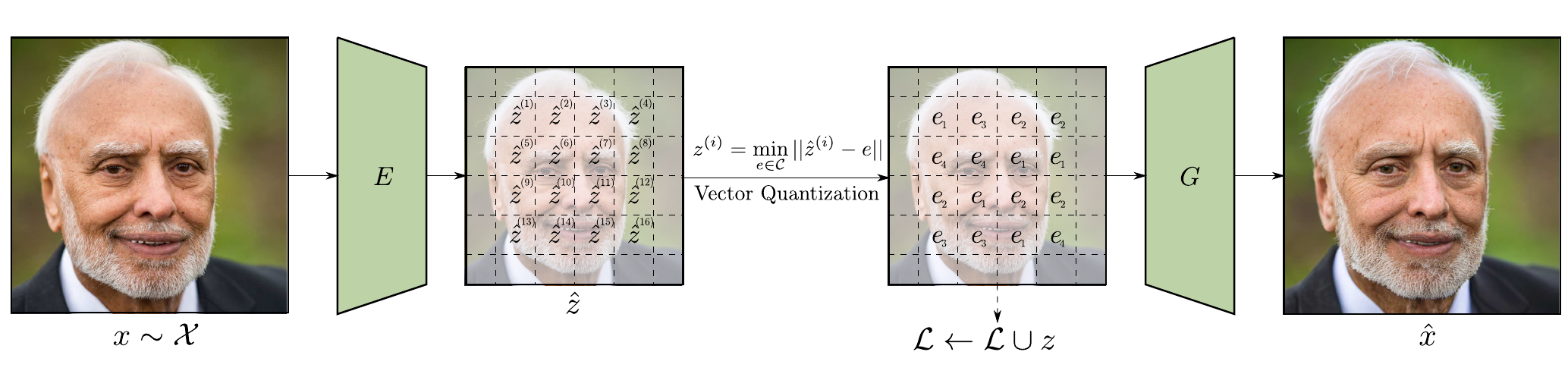}}
    \caption{
        Visualisation of a vector-quantization image model. An encoder model
        extracts continuous representations from the input. Vector quantization
        is then used to map each continuous embedding to the closest entry in
        the codebook, which is then subsequently decoded as a lossy
        reconstruction of the input. We generate a dataset of latent
        representations from an image dataset by iterating $\image \in
        \imageDataset$ and appending the resulting discrete representation
        $\latent$ to a set $\latentDataset$.
    }
    \label{fig:vq}
\end{figure}

Learning useful latent representations, also known as latent codes, in an
unsupervised manner is a key challenge in machine learning. Historically, these
representations have been continuous, but in recent work they are often
discrete. An early example is \gls{vqvae}~\citep{oord2017vqvae}, which has three
main components: an encoder network, a codebook, and a decoder. The encoder
network outputs a compressed continuous representation of the input, and the
codebook $\vqganCodebook$ quantizes these representations, outputting discrete
indices from $1$ to the codebook size $\vqganNbLatents$. Each index $i$ maps to
one of the codebook embeddings (codewords) $e_i$. The decoder maps the quantized
embeddings to a lossy reconstruction of the input. \Gls{vqvae} is trained
end-to-end to reconstruct the input and to minimize the codebook
loss~\citep{oord2017vqvae}.

Once \gls{vqvae} is trained, an auxiliary generative model can be trained to
generate the discrete latent representations. Using \gls{vq} image models helps
to alleviate the computational cost in generative models, as they now operate on
a smaller latent space, rather than in pixel-space. This is especially
significant in \gls{ar} models due to their $\bigO{n}$ iteration complexity.

Due to the rate-distortion trade-off, \gls{vqvae} is not sufficient when the
compression rate $\vqganDownsample$ becomes large, whilst fixing the codebook
size $\vqganNbLatents$. \Gls{vqgan}~\cite{esser2021taming} introduces adversarial and perceptual
loss components to \gls{vqvae}---reformulating the problem as optimising for
perceptual quality rather than minimising distortion. This allows for larger
$\vqganDownsample$ than was possible in prior work before perceptual quality
degrades.

\subsection{Hourglass Transformers}
Vanilla transformers incur a memory complexity of $\bigO{N^2}$ for each
block~\citep{vaswani2017attention}, dominated by costly multi-head
self-attention mechanisms. The majority of research into efficient transformers
focuses on improving the efficiency of the attention mechanisms using sparse
attention patterns~\citep{child2019generating} or through linear complexity
approximations~\citep{xiong2021nystromformer}.

Recent work has focused on making the architecture itself more efficient. Funnel
transformers~\citep{dai2020funneltransformer} progressively downsamples the
input sequence and hence reduces the computational cost of the model. The saved
\glspl{flop} can then be reassigned to create larger models and thus
outperform vanilla transformers given the same computational budget. However,
the final layer does not operate at the same granularity as the input, making it
unusable for tasks such as per-token classification or generative modelling.
Hourglass transformers~\citep{nawrot2021hierarchical} include both up- and
down-sampling mechanisms, resulting in a computational saving whilst being
general-purpose models.

\section{Methodology}
\begin{figure}[ht]
    \centering
    \makebox[\textwidth]{\includegraphics[width=1.1\textwidth]{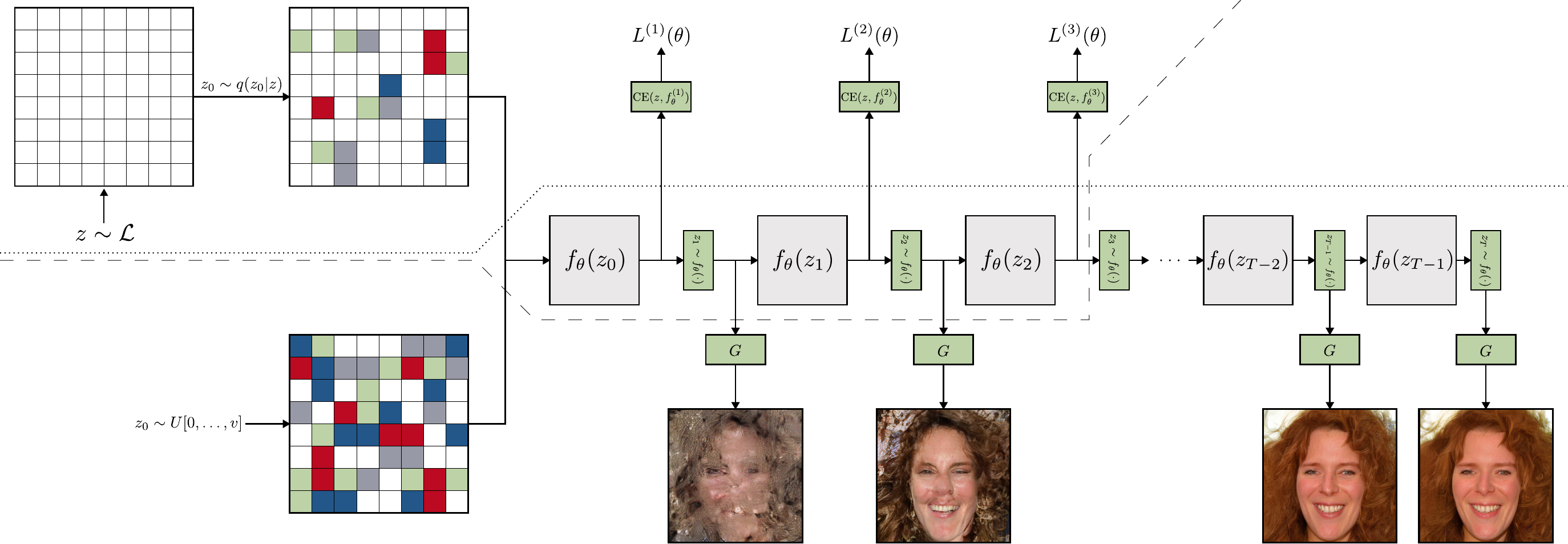}}
    \caption{
        An overview of the \gls{sundae} training and sampling of discrete latent
        representations. Above the dashed line shows the training process,
        whereas below the dotted line shows the sampling process.
        \textbf{Training} begins by sampling $\latent \sim \latentDataset$ and
        then sampling from the corruption distribution $\latent_0 \sim
        q(\latent_0 \vert \latent)$. \Gls{sundae} then denoises for 2 to 3
        steps, calculating the loss at each step. \textbf{Sampling} begins by
        sampling $\latent_0$ from a uniform prior and denoising for
        $\markovSteps$ steps, followed by decoding $\latent_\markovSteps$ to
        obtain $\sample_t$.
    }
    \label{fig:overall}
\end{figure}
Our proposed method aims to push the efficiency of generative models to the
limit via a combination of current techniques. To do so, we first use pretrained
\glspl{vqgan} from~\citet{esser2021taming} to generate a dataset of
discrete latent representations, described in \S\ref{subsec:datasetGen}. By
operating at a latent level, we reduce the spatial resolution for our second
stage generator. We implement our generator as a modified hourglass transformer,
described in \S\ref{subsec:improvedHourglass} and trained using a \gls{nar}
method described in \S\ref{subsec:sundaeTraining}. This permits extremely fast
sampling described in \S\ref{subsec:sundaeSampling}. To thoroughly test the
efficiency and scalability of our approach, we train a megapixel \gls{vqgan}
(described in \S\ref{subsec:megagan}) and repeat \gls{sundae} training and
sampling on the resulting discrete latent representations. In
\S\ref{subsec:inpainting}, we explore flexible inpainting using our framework.
An overview of training and sampling is shown in Figure~\ref{fig:overall}, and
of the latent dataset generation in Figure~\ref{fig:vq}.

Each component represents the pinnacle of performance in their respective area:
compression ratio in \gls{vq} image models with \gls{vqgan}, fast \acrlong{nar}
sampling of discrete data with \gls{sundae}, and transformer scalability with
our modified hourglass transformer. Together, we obtain an extremely efficient
generative model that permits sampling at a resolution of $1024 \times 1024$ in
seconds.

\subsection{Latent Dataset Generation}
\label{subsec:datasetGen}

We use the standard two-stage scheme for \gls{vq} image
modelling~\citep{oord2018neural} using \gls{vqgan}~\citep{esser2021taming} as
our first-stage compression model. For all datasets but FFHQ1024 (see
\S\ref{subsec:megagan}), we use pretrained \glspl{vqgan}.

The second stage is to train a discrete prior model over the extracted latent
representations. To enable this, we generated a latent dataset using a trained
\gls{vqgan}. This allows for faster training of our discrete prior as the
latent representations have been precomputed. A downside of this
approach is that it limits the amount of data augmentation that can be applied
to the dataset. We apply a simple horizontal flip to all images, effectively
doubling the dataset size, with no other augmentation. Formally, given a dataset
of images $\imageDataset$, a \gls{vqgan} encoder $\vqganEncoder$ with downsample
factor $\vqganDownsample$, and \gls{vq} codebook $\vqganCodebook$ with number of
codewords $\vqganNbLatents$, trained on $\imageDataset$, we define our latent
dataset $\latentDataset$ as:
\begin{equation}
    \latentDataset = \{\vqganCodebook(\vqganEncoder(\image)) \mid \image \in \imageDataset \}
\end{equation}
where $\image \in \real{3 \times H \times W}$ is a single element of the
augmented image
dataset $\imageDataset$ and $\latent = \vqganCodebook(\vqganEncoder(\image)) \in \{1, \dots,
\vqganNbLatents\}^{h \times w}$ is the corresponding discrete latent
representation. In other words, each $\vqganDownsample \times \vqganDownsample$
patch in $\image$ is mapped to a single discrete value from $1$ to
$\vqganNbLatents$ (corresponding to a codeword $\codebookVector \in
\vqganCodebook$),
resulting in a latent representation of shape $\frac{H}{f} \times \frac{W}{f} =
h \times w$.

We then use $\latentDataset$ to train a discrete prior over the latents. Coupled
with the \gls{vqgan} decoder $\vqganDecoder$, we obtain a powerful generative
model by first sampling $\latent_0$ from a uniform prior distribution,
iteratively denoising using \gls{sundae}, and then decoding
$\latent_\markovSteps$ using the \gls{vqgan} decoder $\vqganDecoder$ to obtain
the final sample $\sample$. The training of this discrete prior model forms the
bulk of our work in this paper.

\subsection{Multidimensional Hourglass Transformer}
\label{subsec:improvedHourglass}
\begin{figure}[htp]
    \centering
    \makebox[\textwidth]{\includegraphics[width=1.2\textwidth]{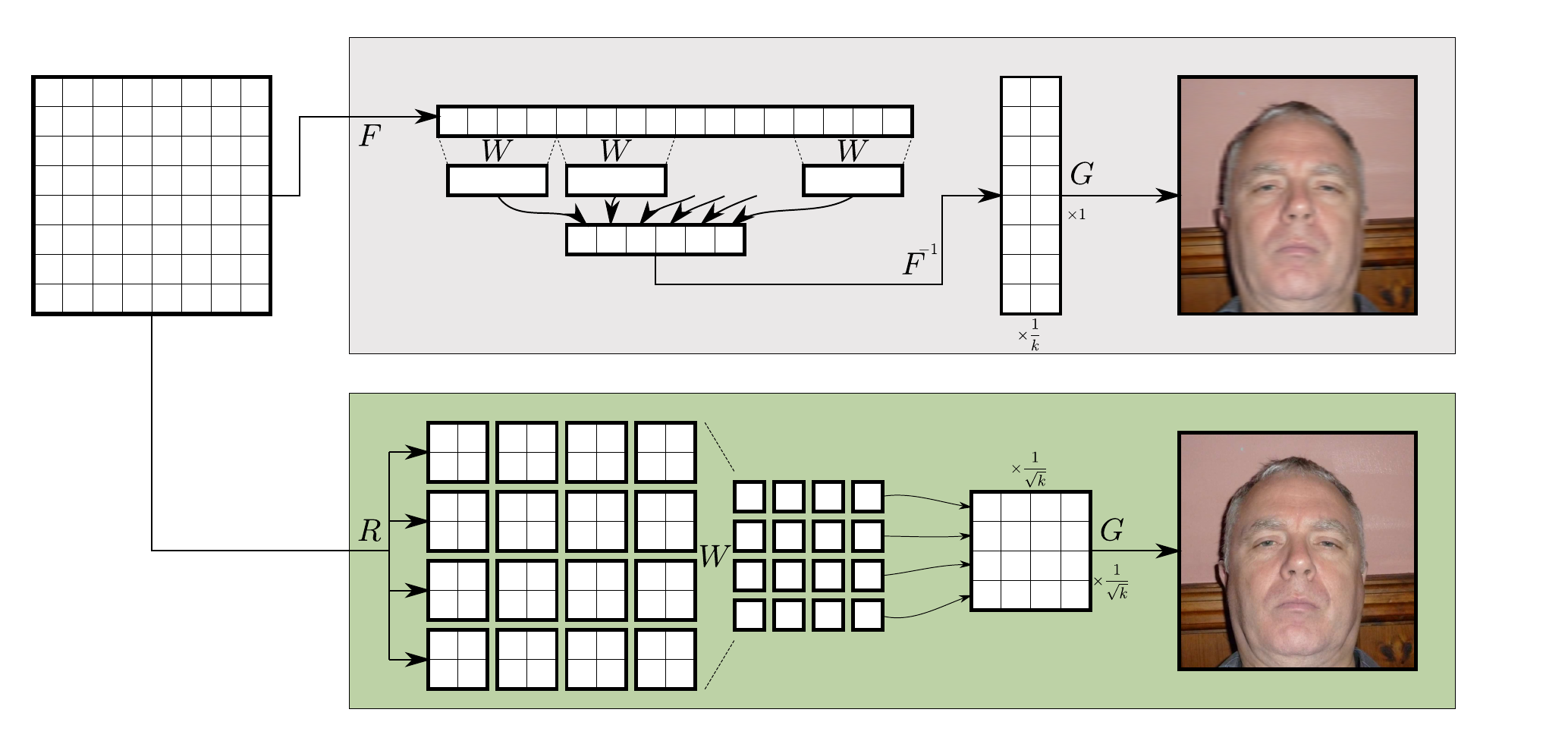}}
    \caption{
        \textbf{Top:} Showing the effect of resampling sequence embeddings using
        original formulation. Resampling will be applied to only one axis,
        resulting in resampling in only one axis of the decoded image.
        \textbf{Bottom:} Our method of resampling, extracting 
        two-dimensional patches of size $\sqrt{\hourglassRate}$, then 
        resampling. The sequence is then flattened and passed to
        subsequent transformer layers.
    }
    \label{fig:resample}
\end{figure}

Inspired by successes in hierarchical transformers for language
modelling~\citep{nawrot2021hierarchical}, we chose to apply it to the task of
discrete latent modelling. Hourglass transformers have been shown to efficiently
handle long sequences, outperform existing models using the same computational
budget, and meet the same performance as existing models more efficiently by
utilising a hierarchical structure~\citep{nawrot2021hierarchical}. The same
benefits should also apply to \gls{vq} image modelling. 

However, the design and parameters chosen by the original authors are tailored
for language modelling~\citep{nawrot2021hierarchical} with limited experiments
on image generation (channel-wise downsampling only). We improve upon their
architecture for our task of discrete latent modelling, which we believe may
also be applicable to multidimensional inputs generally. We leave confirmation
of this to future work, and outline modifications in this subsection.

\textbf{2D-Aware Downsampling}---The original formulation of hourglass
transformers~\citep{nawrot2021hierarchical} introduced up- and down-sampling
layers, allowing the use of hierarchical transformers in tasks that have equal
input and output sequence lengths. However, we found certain flaws in their
original formulation that hinders performance on multidimensional inputs.

In their work, resampling is applied to the flattened embedding sequences,
meaning that a corresponding two-dimensional vector-quantized image is resampled
more in one axis compared to the other. In their work this was not addressed, as
multidimensional experiments were limited to channel-wise sampling with no
spatial resampling~\citep{nawrot2021hierarchical}. 

Instead, we reshape the flattened sequence into a two-dimensional form and apply
resampling equally in the last two axes. With a resampling rate of
$\hourglassRate$ we apply $\sqrt{\hourglassRate}$ in each spatial axis---evenly
distributing the resampling among the axes. In our preliminary experiments, this
significantly improved the performance of the discrete prior model. A comparison
between the previous approach and our amended solution is shown in
Figure~\ref{fig:resample}.

For our resampling operations, we use linear resampling (following
recommendation by \citet{nawrot2021hierarchical} to use linear resampling for
image tasks) and a post-resampling attention layer, providing global context and
aggregation of information to the resampling operations. Our
adjusted resampling method is as follows:
\begin{equation}
    h' = A(\hourglassResample^{(\intercal)} \cdot R(\vh) + \vr), \;\; \hourglassResample \in
    \mathbb{R}^{\frac{(d \cdot h \cdot w)}{\hourglassRate} \times (d \cdot h \cdot w)}
\end{equation}
where $A$ is the post-resampling attention layer, $\vh$ is the hidden
state of size $d$, $\vr$ is the residual (with $\vr = \bm{0}$ when downsampling), $R$ is the
modified reshape operation, and $\textbf{W}$ is a
learned projection matrix. The reshape operation $R$ was implemented as a
space-to-depth operation followed by combining the feature and depth dimensions.

\textbf{Rotary Positional Embeddings}---Transformers have no inductive biases
that allow it inherently know the position of an element in the sequence.
Embeddings that represent positions must be injected into the model in addition
to the input itself. In our work, we choose to use rotary positional
embeddings~\citep{su2021roformer} as they require no additional parameters, incur
only a small runtime cost, and can be easily extended to the multidimensional
case~\citep{rope-eleutherai}, which we exploit here. Though transformers are
clearly capable of learning that elements far apart in a flattened sequence may
be semantically close, we found that explicitly extending positional embeddings
to the multidimensional case to provide a modest boost in performance and
improve the rate of training convergence. The original hourglass transformer on
pixel-wise generation also opted to use rotary
embeddings~\citep{nawrot2021hierarchical} but in the single dimensional case.
Though rotary embeddings are agnostic to the self-attention method used, we use
full self-attention~\cite{vaswani2017attention} in all experiments.

\textbf{Removal of Causal Constraints}---In the original \gls{ar}
formulation of hourglass transformers they noted that naively resampling could
cause information leakage into future sequence elements---therefore violating
the autoregressive property~\citep{nawrot2021hierarchical}. As our approach is
\gls{nar} we do not make any special considerations to avoid information leaking
into the future. This simplifies the model by avoiding shifting and causal
masking operations required in the original work.

\subsection{Non-autoregressive Discrete Prior Training}
\label{subsec:sundaeTraining}
We train a \gls{sundae} model on the flattened (in a raster-scan format)
\gls{vq} latents $\latent = \{\latent^{(0)}, \dots, \latent^{(N)}\}$
where $N = \latentHeight \cdot \latentWidth$. The function $\sundae(\cdot)$ is
implemented using a multidimensional hourglass transformer. 

Given a latent $\latent \sim \latentDataset$, we apply our corruption
distribution. This is done by first sampling a corruption threshold vector
$\corruptionThreshold$ with $\corruptionThreshold_i \sim U[0, 1]$ and a random
matrix $\mathbf{R}$ of the same shape as $\latent$ where $R_{i,j} \sim U[0,1]$.
Using this, we construct a mask matrix $\mathbf{M}$ with $M_{i,j} = 1$ when
$R_{i,j} < t_i$ and $M_{i,j} = 0$ otherwise. This results in $\mathbf{M}_i$ having
approximately $\corruptionThreshold_i$ of its entries be $1$.

Then, given $\latent_0 \sim p_0$, we update the $\latent_0$ to start unrolled
denoising from:
\begin{equation}
    \latent_0 \leftarrow \mathbf{M} \cdot \latent_0 + (\mathbf{1} - \mathbf{M})
    \cdot \latent \text{.}
\end{equation}

We then iteratively unroll the current sample $\latent_{t-1}$ to obtain
$\latent_t$ for steps $t\in \{1, \dots, \markovSteps\}$. To perform one unroll
step, simply compute logits $\sundae(\latent_t \vert \latent_{t-1})$ and then
sample from the resulting distribution to obtain $\latent_t$, storing the logits
at each step $t$. Then, compute the cross entropy loss between all logits at each
$t$ and the target $\latent$. This differs from some \gls{nar} solutions
which predict the corruption noise~$\epsilon$~\citep{ho2020ddpm} rather than the
target. The mean of the cross entropy losses is then computed to produce
the final loss: 
\begin{equation} 
    \lossFunction{1:T} = \frac{1}{T} \left(\lossFunction{1} +
    \dots + \lossFunction{T} \right) 
\end{equation} 
as in \S\ref{subsec:sundae}, which allows for the backpropagation of gradients
and consequently the updating of parameters $\sundaeParameters$, with
$\markovSteps=2$. 

An alternative corruption distribution would be to instead use a deterministic
method $\latent_0^{(i)}=\texttt{[MASK]}=\vqganNbLatents+1$, essentially
replacing all tokens with $M_{i,j} = 1$ with a special masking token. This is
similar to ``progressive unmasking'' of latents used in prior
work~\citep{bondtaylor2021unleashing}. This strategy was not
considered as the use of a masking token places an upper bound on $\markovSteps$
during sampling (updating at most one token per step) as well as not allowing
for self-correction, as once a token is unmasked it becomes
fixed.

\subsection{Fast Image Generation}
\label{subsec:sundaeSampling}
During sampling, we sample $\latent_t \sim \sundae(\latent_t \vert
\latent_{t-1})$ for a constant number of steps $\markovSteps$, beginning
randomly from $\latent_0$ sampled from a uniform
distribution~\citep{savinov2022stepunrolled}. The original work proposed a number
of improved strategies for sampling in a smaller number of steps, including
low-temperature sampling and updating a random subset of
tokens rather than all simultaneously.

Sampling with a lower temperature, however, reduces the diversity of the
outputs. To alleviate this, we anneal the temperature down from a high value
($\approx 1.0$) down to a low value during the sampling process. We found this
retained the fast sampling speed whilst improving diversity.

In certain latent sampling configurations, updating only a small subset of
tokens also helps improve diversity. \citet{savinov2022stepunrolled} used this
strategy when performing low-temperature sampling. However, we found that for
low-step sampling ($\markovSteps\leq 50$) that a lower sample proportion meant
not all tokens are updated enough, resulting in poor quality samples. Hence, in
these cases, we do not follow their proposal and instead use a high sample
proportion ($0.5\sim 0.8$). In scenarios where we are permitted a time-budget
allowing for many sampling steps, the sample proportion can be freely reduced
for an increase in sample diversity.

If an individual sample does not change between step $t-1$ and $t$,
it is prevented from being changed further. If all samples are frozen, sampling
terminates early, provided a minimum number of steps have been completed. This
improves the sampling speed with little cost to the sample quality. This
is significant when performing large-batch sampling or when $\markovSteps$ is
large.

Once sampling has terminated, the sampled latent code $\latent_\markovSteps$ is
decoded by the \gls{vqgan} decoder $\vqganDecoder$ to produce a final sample
$\sample$. In fact, any $\latent_i$ in the Markov chain is a valid input to
$\vqganDecoder$. Decoding during sampling and visualising $\sample_t$ each step
$t$ shows the model gradually denoising the latent and correcting errors
accumulated during sampling.

\subsection{Scaling VQ-GAN to Megapixel Images}
\label{subsec:megagan}
\begin{figure}[ht]
    \centering
    \makebox[\textwidth]{\includegraphics[width=1.1\textwidth]{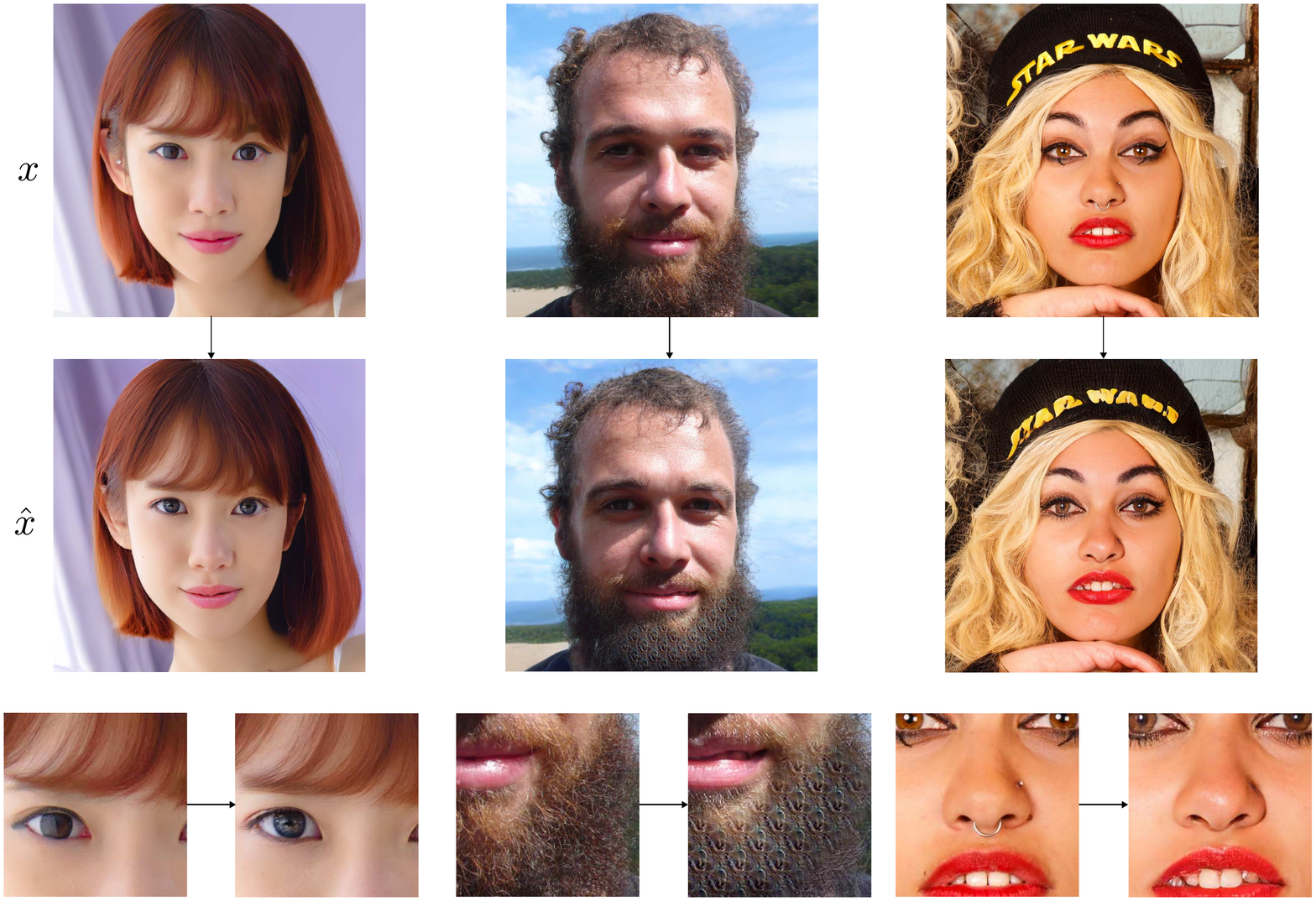}}
    \caption{
        \Gls{vqgan} does not always produce faithful reconstructions due to
        being optimised for perceptual quality rather minimising distortion.
        \textbf{Left}: The eye colour has been brightened; hair shifted to
        conceal an ear piercing, rather than reconstruct it.
        \textbf{Middle}: Mode collapse in hair texture; the pose of the lip is
        altered. 
        \textbf{Right}: Text in image is corrupted; nose and lip piercings are
    removed; eye makeup is altered.}
    \label{fig:recon}
\end{figure}

Training at higher resolutions means greater computational requirements and
slower sampling speeds. With an \gls{ar} model, the sampling time can be immense
as it scales linearly with data dimensionality, even with an auxiliary \gls{vq}
image model~\citep{esser2021taming}. With a \gls{nar} model however, the
sampling speed is explicitly controlled and does not directly grow as a function
of input size, meaning our proposed generative framework is highly scaleable. To
demonstrate this, we applied our proposed framework to megapixel image
generation.

We trained a larger variant of \gls{vqgan} with $\vqganNbLatents = 8192$
operating on $1024 \times 1024$ from FFHQ1024. To our knowledge, this is the
highest resolution dataset \gls{vqgan} has been applied to. Once trained, we
generate the latent dataset as before, the only difference being an increased
sequence length---greater than \gls{sundae} was ever tested on in the original
work~\citep{savinov2022stepunrolled}. Specifically, we obtain a downsampling
rate of $\vqganDownsample=32$, resulting in discrete latents of size $32 \times
32 = 1024$.

The resulting reconstructions are overall of good quality given the large
compression ratio in use. However, the reconstructions are not without
artifacts. Figure \ref{fig:recon} shows examples of particularly prevalent
artifacts including occasional unrealistic textures in hair and corruption of
text. The corruption of text is a common issue in \gls{vq} image
models~\citep{ramesh2021dalle}, and the unrealistic textures are likely a result
of a lack of model capacity or patch-wise mode collapse.

\Gls{vqgan} is trained to minimise the mean absolute error, perceptual loss, and 
adversarial loss~\citep{esser2021taming} in addition to a $k$-means \gls{vq}
loss. Specifically, \gls{vqgan} is trained to minimise the following
loss:
\begin{align}
\begin{split}
    L_\text{PIX} &= \alpha_\text{PIX} \cdot |\image - \hat{\image}| \cdot \\
    L_\text{VQ} &= \alpha_\text{VQ} \cdot \left(||\hat{\latent} -
    \latent||^2 + ||sg[\vqganEncoder(x)] - \latent||^2_2 + ||\vqganEncoder(x) -
    sg[\latent]||^2_2\right)\\
    L_\text{GAN} &= \alpha_\text{GAN} \cdot \left(\log D(x) + \log
    (1-D(\hat{x}))\right) \\
    \lambda &= \frac{\nabla_{G_{-1}}[L_\text{PIX} +
    L_\text{PER}]}{\nabla_{G_{-1}}[L_\text{GAN}] + \epsilon}\\
    L &= L_\text{VQ} + \lambda \cdot L_\text{GAN}\\
    \alpha_\text{PIX} &= 1.0,\; \alpha_\text{VQ} = 1.0,\; \alpha_\text{GAN} = 0.5,\; \alpha_\text{PER} = 1.0
\end{split}
\label{eq:vqgan}
\end{align}
where $\nabla_{G_{-1}}[\cdot]$ is the gradient with
respect to the last layer of the \gls{vqgan} decoder $\vqganDecoder$ and
$sg[.]$ is the stop-gradient operator. The
generator and discriminator model parameters are updated separately, as is
standard procedure in \gls{gan}-based literature~\citep{esser2021taming}.

As there is less relative weight in Equation~\ref{eq:vqgan} on minimising
distortion, this gives rise to an interesting property of \gls{vqgan} where the
reconstructions may be perceptually valid but distinct from the input. The left
reconstruction in Figure \ref{fig:recon} demonstrates this with a change in eye
colour and a shift in hair position---concealing an ear-piercing. This even
more apparent in the right reconstruction where all piercings are flawlessly
removed---along with adjustments to eye makeup.

Using \gls{vq} image models to compress images further whilst retaining high
quality and faithful reconstructions remains an open and challenging area of
research---particularly at high resolutions. In our preliminary
experiments, we found a higher $\vqganDownsample$ led to the majority of
reconstructions being of an untenable quality. Conversely, decreasing
$\vqganDownsample$ led to latent representations of sizes that resulted in large
memory requirements in the downstream \gls{sundae} prior, making inference on
consumer-grade GPUs impractical.

Training \gls{vqgan} at this resolution and at our chosen downsampling rate is
extremely computationally expensive. This made a full hyperparameter sweep of
\gls{vqgan}'s parameters not possible. Therefore, we accepted good
reconstructions on average with occasional artifacts that could potentially
manifest in the final samples. Improving the effectiveness of \gls{vq} image
models is not the focus of this research project. We found these artifacts to
only rarely appear in the final samples, shown in
\S\ref{subsec:evaluationUnconditional}.

\subsection{Arbitrary Pattern Inpainting}
\label{subsec:inpainting}
As noted in the original work~\citep{savinov2022stepunrolled} and other \gls{nar}
solutions~\citep{bondtaylor2021unleashing}, one advantage of \gls{nar} models is
that they are not limited to causal inpainting. They support arbitrary
inpainting masks and draw upon context from $\latent_{t-1}$, rather than
$\latent_{t-1}^{<i}$, enabling them to easily perform inpainting tasks that are
complex to implement with \gls{ar} models. This property also results in 
higher quality and more diverse samples~\citep{bondtaylor2021unleashing}.

The inpainting procedure takes an image $\sample \in \real{H \times W \times 3}$
and a pixel-level binary mask $\pixelMask \in \{0, 1\}^{H \times W}$ as input.
By taking $\vqganDownsample \times \vqganDownsample$ regions of $\pixelMask$ and
applying a logical \texttt{AND} in them, we obtain a latent level mask $\vqMask
\in \{0,1\}^{h \times w}$. We encode $\image$ using $\vqganEncoder$ to obtain
$\latent$, and then initialise our starting latent $\latent_0$ by randomly
setting points in $\latent$ where $m_\text{vq} = 1$, and keeping $\latent$ the
same when $m_\text{vq} = 0$. We then sample starting from $\latent_0$, allowing
the model full context, but only update regions that were masked according to
$\vqMask$. Though not strictly necessary, we use a lower temperature for
inpainting ($0.3 \leq \temperature \leq 0.5$) to bias sampling towards more
confident choices, as more true context is available to the model. Decoding
$\latent_T$ with $\vqganDecoder$ produces the final result $\sample$, identical
to the end of the sampling process.

Sampling at a latent level means the model is unable to do fine-grained
inpainting at a pixel level. The definition of the \gls{vq} mask $\vqMask$ means
that some pixels outside the mask may be altered if the pixel mask is not
perfectly aligned with the \gls{vq} mask (when \texttt{AND} does not always
receive all 0s or all 1s). We found in practise this had little effect on the
perceptual quality of the outputs.

\section{Evaluation}
\subsection{Unconditional Image Generation}
\label{subsec:evaluationUnconditional}
\begin{figure}[ht]
    \centering
    \makebox[\textwidth]{\includegraphics[width=1.1\textwidth]{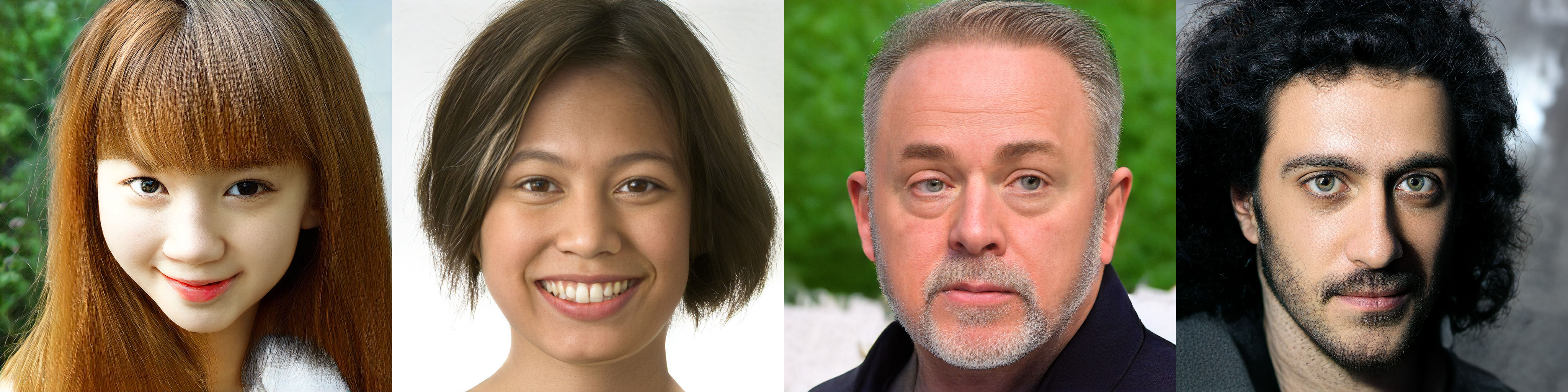}}
    \caption{
        Unconditional FFHQ $1024 \times 1024$ samples generated
        in $\markovSteps = 100$ sampling steps.
    }
    \label{fig:face}
\end{figure}

\begin{table}[h!]
\centering
\begin{tabular}{ |c|c|c|c| } 
    \hline
    \textbf{Model} & \textbf{FID} & \textbf{Density} & \textbf{Coverage} \\
    \hline
    \gls{vqgan} + Transformer~\citep{esser2021taming} & 9.76 &---& ---\\ 
    \gls{vqgan} + Absorbing DDPM~\cite{bondtaylor2021unleashing} & 6.11 & 1.51 & 0.83\\ 
    ViT-VQGAN + Transformer~\citep{yu2021vqgan} & 5.3 &---& ---\\ 
    \hline
    \textbf{\gls{vqgan} + SUNDAE} ($\markovSteps = 200$) & 11.79 & 0.72 & 0.21 \\
    \textbf{\gls{vqgan} + SUNDAE} ($\markovSteps = 100$) & 10.56 & 0.77 & 0.43 \\
    \hline
\end{tabular}
\caption{Results on FFHQ $256 \times 256$ for unconditional image generation.}
\label{tab:ffhq}
\end{table}

\begin{table}[h!]
\centering
\begin{tabular}{ |c|c|c|c| } 
    \hline
    \textbf{Model} & \textbf{FID} & \textbf{Density} & \textbf{Coverage} \\
    \hline
    \gls{vqgan} + Transformer~\citep{esser2021taming} & 10.2 &---& ---\\ 
    ViT-VQGAN + Transformer~\citep{yu2021vqgan} & 7.0 &---& ---\\ 
    \hline
    \textbf{\gls{vqgan} + SUNDAE} ($\markovSteps = 200$) & 21.25 & 0.40 & 0.33 \\
    \textbf{\gls{vqgan} + SUNDAE} ($\markovSteps = 100$) & 17.96 & 0.43 & 0.36 \\
    \hline
\end{tabular}
\caption{Results on CelebA $256 \times 256$ for unconditional image generation.}
\label{tab:celeba}
\end{table}

We evaluate our method on the task of unconditional image generation using datasets
FFHQ256, FFHQ1024, and CelebA. We evaluate our models using FID
Infinity~\citep{chong2020effectively}, coverage, and
density~\citep{ferjad2020icml}. Additionally, we show representative
unconditional samples in Figures~\ref{fig:main} \& \ref{fig:face}, with
additional samples in Appendix~\ref{sec::additionalFFHQ}.

We present our results for the $256 \times 256$ image generation experiments in
Table~\ref{tab:ffhq} and Table~\ref{tab:celeba} for FFHQ and CelebA
respectively. Though the samples for models trained on these datasets are of
good quality, they do not improve upon prior work utilising \gls{vqgan} to
generate images, in terms of our chosen perceptual quality metrics. However, our
FID score for FFHQ256 falls close to the reported score in
\citet{esser2021taming} using less than half the number of sampling steps.
Additionally, both \citet{bondtaylor2021unleashing} and \citet{yu2021vqgan}
utilise improved \glspl{vqgan}, whereas we use the original formulation. This
suggests that applying \gls{sundae} to these variants could improve our scores
further. 

For the megapixel experiments, samples shown in Figures~\ref{fig:main} \&
\ref{fig:face} demonstrate that our model is capable of generating high quality
and diverse samples. Our aim was to push the efficiency of generative models to
their limit, however we were still surprised at precisely how fast the model
could sample---particularly on megapixel scale experiments. The samples in
Figure~\ref{fig:main} were created in two seconds on a GTX 1080Ti. This can be
further improved with more powerful accelerators and further optimisation.
Additionally, our success here demonstrates the scalability of \gls{sundae} to
sequence lengths of $N=1024$, larger than the maximum length tested in the
original work ($N=256$) that proposed
\gls{sundae}~\citep{savinov2022stepunrolled}.

\begin{figure}[htb!]
    \begin{subfigure}[b]{0.32\textwidth}
        \centering
        \resizebox{\textwidth}{!}{
            \begin{tikzpicture}
\begin{axis}[
y label style={at={(axis description cs:-0.15,1.0)},rotate=-90,anchor=south},
title={},
xlabel={Sampling proportion},
ylabel={FID $\downarrow$},
xmin=0.3, xmax=0.9000000000000001,
ymin=0, ymax=120,
xtick={0.3,0.4,0.5,0.6000000000000001,0.7000000000000002,0.8000000000000003,0.9000000000000001},
ytick={0,20,40,60,80,100,120},
legend pos=north east,
ymajorgrids=true,
grid style=dashed,
]\addplot[color=black, mark=square]
coordinates {(1.0, 70.67575660798256)(0.8, 47.19247310169379)(0.6, 27.254076458757616)(0.4, 19.97191697142088)(0.2, 13.348316234479821)};
\addlegendentry{FFHQ256}
\addplot[color=red, mark=*]
coordinates {(1.0, 100.01221541159126)(0.8, 60.19022147026466)(0.6, 28.470073088341294)(0.4, 21.87382632148365)(0.2, 22.2484863764138)};
\addlegendentry{FFHQ1024}
\addplot[color=blue, mark=diamond]
coordinates {(1.0, 97.83524644278657)(0.8, 60.84783349797591)(0.6, 40.04168819808587)(0.4, 29.745576875131956)(0.2, 21.158200179625435)};
\addlegendentry{CelebA}
\end{axis}
\end{tikzpicture}
        }
        \caption{Proportion vs. FID.}
    \end{subfigure}
    \begin{subfigure}[b]{0.32\textwidth}
        \centering
        \resizebox{\textwidth}{!}{
            \begin{tikzpicture}
\begin{axis}[
y label style={at={(axis description cs:-0.15,1.0)},rotate=-90,anchor=south},
title={},
xlabel={Sampling proportion},
ylabel={Coverage $\uparrow$},
xmin=0.3, xmax=0.9000000000000001,
ymin=0.0, ymax=1.8,
xtick={0.3,0.4,0.5,0.6000000000000001,0.7000000000000002,0.8000000000000003,0.9000000000000001},
ytick={0.0,0.2,0.4,0.6000000000000001,0.8,1.0,1.2000000000000002,1.4000000000000001,1.6,1.8},
legend pos=north east,
ymajorgrids=true,
grid style=dashed,
]\addplot[color=black, mark=square]
coordinates {(1.0, 0.2754)(0.8, 0.4407)(0.6, 0.6247116968698517)(0.4, 0.649)(0.2, 0.7398)};
\addlegendentry{FFHQ256}
\addplot[color=red, mark=*]
coordinates {(1.0, 0.0934)(0.8, 0.2186)(0.6, 0.4145)(0.4, 0.473)(0.2, 0.5156)};
\addlegendentry{FFHQ1024}
\addplot[color=blue, mark=diamond]
coordinates {(1.0, 0.1198)(0.8, 0.22)(0.6, 0.2864)(0.4, 0.3267)(0.2, 0.3868)};
\addlegendentry{CelebA}
\end{axis}
\end{tikzpicture}
        }
        \caption{Proportion vs. Coverage.}
    \end{subfigure}
    \begin{subfigure}[b]{0.32\textwidth}
        \centering
        \resizebox{\textwidth}{!}{
            \begin{tikzpicture}
\begin{axis}[
y label style={at={(axis description cs:-0.15,1.0)},rotate=-90,anchor=south},
title={},
xlabel={Sampling proportion},
ylabel={Density $\uparrow$},
xmin=0.3, xmax=0.9000000000000001,
ymin=0.0, ymax=1.8,
xtick={0.3,0.4,0.5,0.6000000000000001,0.7000000000000002,0.8000000000000003,0.9000000000000001},
ytick={0.0,0.2,0.4,0.6000000000000001,0.8,1.0,1.2000000000000002,1.4000000000000001,1.6,1.8},
legend pos=north east,
ymajorgrids=true,
grid style=dashed,
]\addplot[color=black, mark=square]
coordinates {(1.0, 1.2679999999999998)(0.8, 0.9545666666666667)(0.6, 0.9844041735310269)(0.4, 0.9652333333333334)(0.2, 1.1701333333333332)};
\addlegendentry{FFHQ256}
\addplot[color=red, mark=*]
coordinates {(1.0, 0.35050000000000003)(0.8, 0.4302333333333333)(0.6, 0.5753333333333333)(0.4, 0.7180666666666666)(0.2, 0.8203333333333332)};
\addlegendentry{FFHQ1024}
\addplot[color=blue, mark=diamond]
coordinates {(1.0, 0.6190666666666667)(0.8, 0.6029333333333333)(0.6, 0.4970333333333333)(0.4, 0.4465666666666666)(0.2, 0.45296666666666663)};
\addlegendentry{CelebA}
\end{axis}
\end{tikzpicture}
        }
        \caption{Proportion vs. Density.}
    \end{subfigure}
    \caption{
        Plots showing sample quality in terms of different perceptual metrics as
        sample proportion is changed. Lower proportions perform better provided
        $\markovSteps$ is sufficiently high.
    }
    \label{fig:prop}
\end{figure}
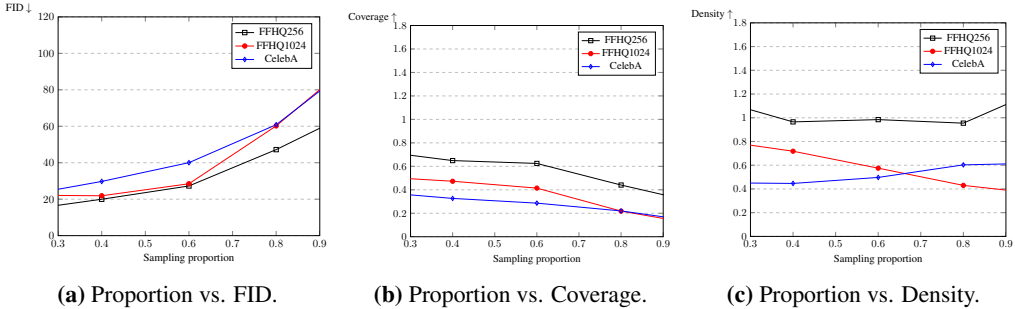

However, in terms of perceptual quality metrics, our approach again does not
improve upon prior work, obtaining a FID score of $21.85$, density of $0.77$,
and coverage of $0.47$ using $\markovSteps = 100$ sampling steps. This is likely
due to being bottlenecked by our relatively weak \gls{vqgan} prior model.
Further work is needed to improve the reconstruction quality of \gls{vqgan} on
images at this resolution. It should be emphasized that an \gls{ar} model
learning the same FFHQ1024 prior would require $\markovSteps~=~1024$ sampling
steps. Additionally, across all datasets, reasonable samples can be obtained in
as few as $\markovSteps = 50$ sampling steps, such as those in
Figure~\ref{fig:main}, though sample quality is more inconsistent.

Table~\ref{fig:prop} shows that recommendations to prefer low proportion
sampling not only holds for natural language tasks
\citep{savinov2022stepunrolled} but also for the sampling of discrete latents.
Across all datasets, the best FID and coverage scores were obtained with our
minimum tested proportion of $0.2$. It should be noted that this is only
possible when a sufficient number of sampling steps is available to the model,
else each individual position in $\latent_t$ will not have enough opportunities
to update. Additionally, Table~\ref{tab:ffhq} and Table~\ref{tab:celeba} show
that naively increasing the number of sampling steps (without changing other
parameters) does not necessarily lead to improved performance. Further work is
needed to study best practises for adjusting other sample-time parameters when
scaling the number of sampling steps.

\subsection{Class-conditioned Image Generation}
\begin{figure}[ht]
    \centering
    \begin{subfigure}[b]{0.47\textwidth}
        \centering
        \includegraphics[width=1.0\textwidth]{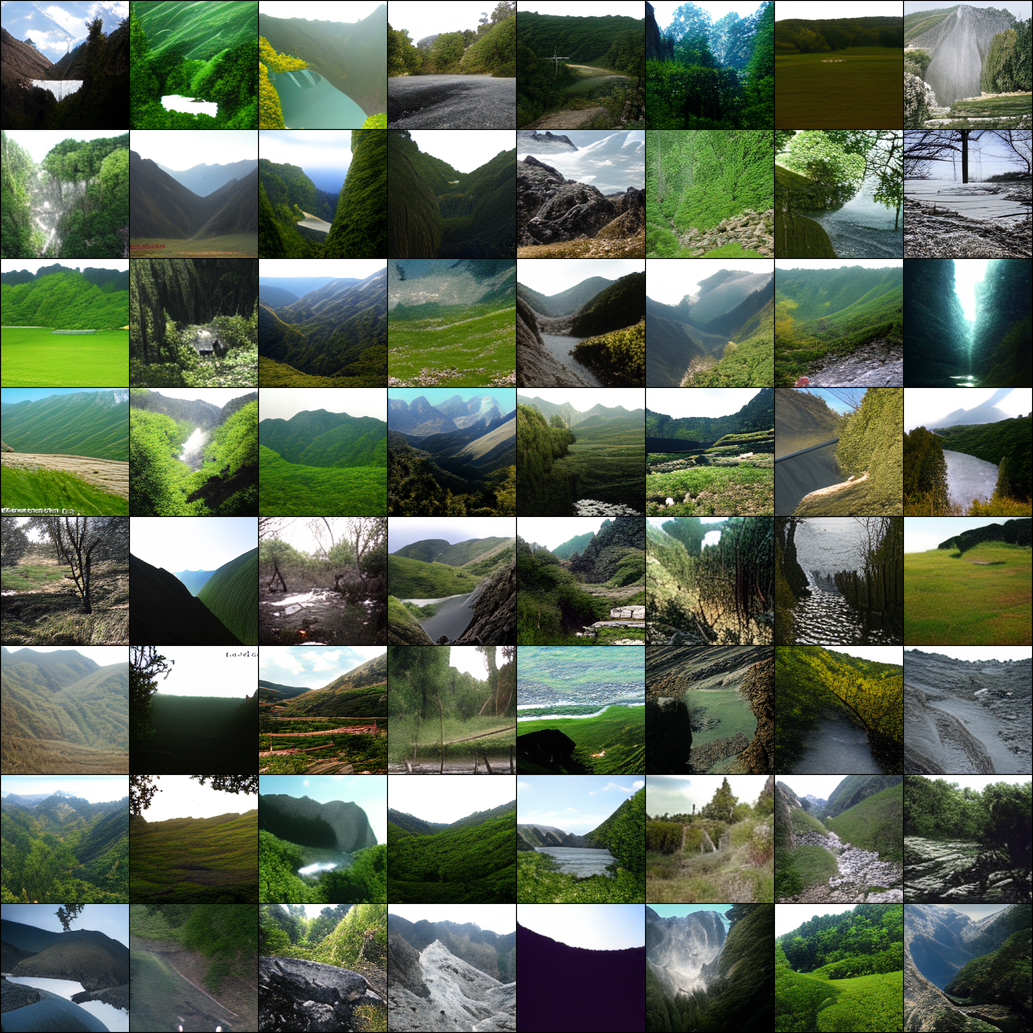}
        \caption{
            $256 \times 256$ successful samples from the class ``Valley''.}
    \end{subfigure}
    \hfill
    \begin{subfigure}[b]{0.47\textwidth}
        \centering
        \includegraphics[width=1.0\textwidth]{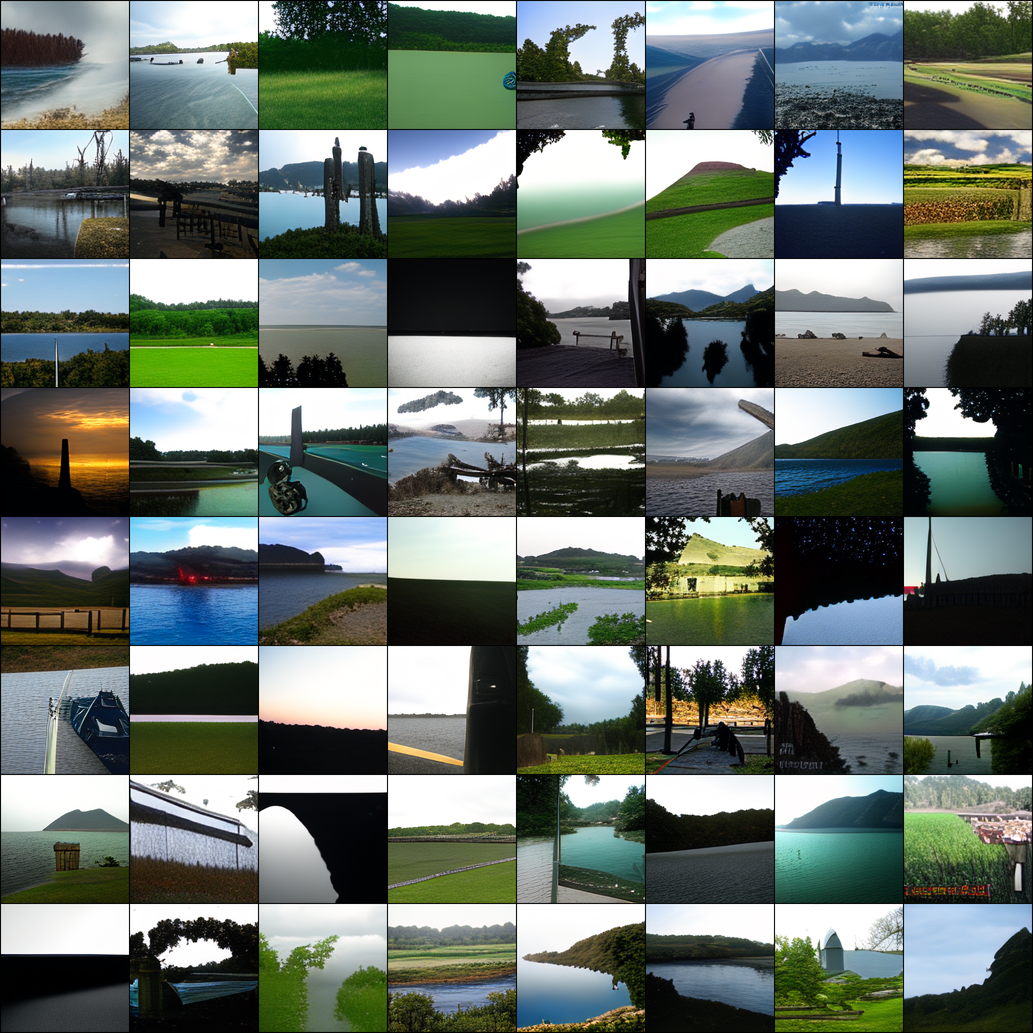}
        \caption{
            $256 \times 256$ successful samples from the class ``Lakeside''.
        }
    \end{subfigure}\\
    \vspace{0.5cm}
    \begin{subfigure}[b]{0.47\textwidth}
        \centering
        \includegraphics[width=1.0\textwidth]{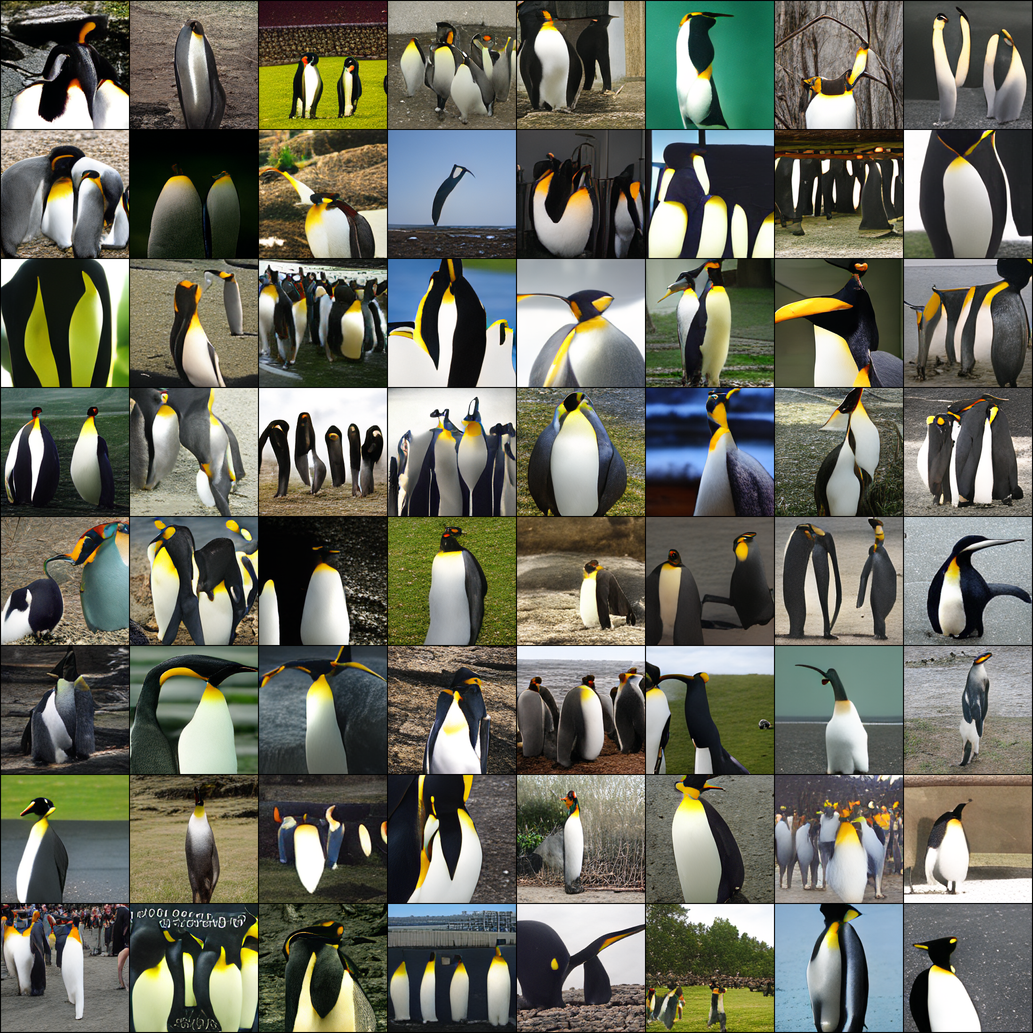}
        \caption{
            $256 \times 256$ failed samples from the class ``King Penguin''.
        }
    \end{subfigure}
    \hfill
    \begin{subfigure}[b]{0.47\textwidth}
        \centering
        \includegraphics[width=1.0\textwidth]{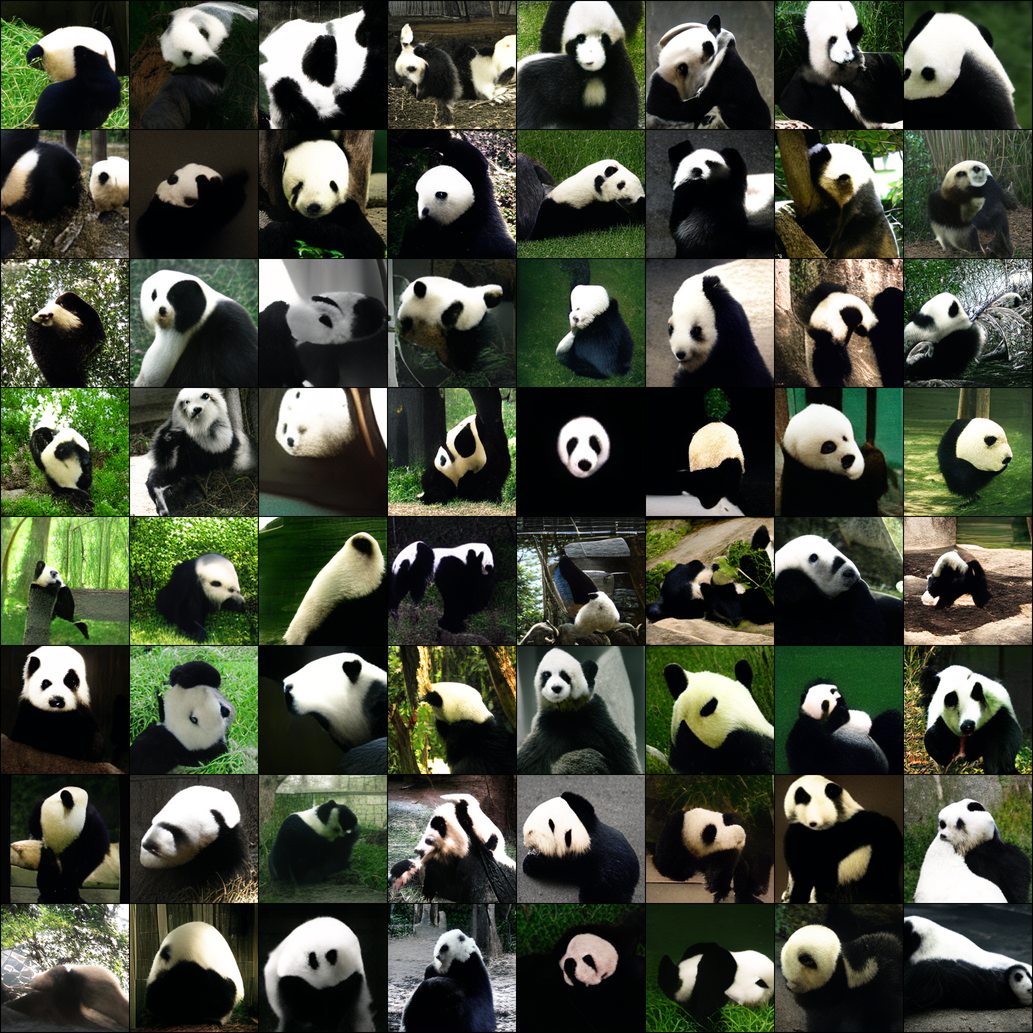}
        \caption{
            $256 \times 256$ failed samples from the class ``Giant Panda''.
        }
    \end{subfigure}
    \caption{
        Examples of class-conditioned generation on ImageNet256 using
        $\markovSteps = 50$ sampling steps. Top row contains examples of
        successful samples whereas bottom row contains failed samples. The
        contents of the failed samples resemble the target class, but are of low
        quality.
    }
    \label{fig:imagenet}
\end{figure}

\begin{figure}[ht]
    \centering
    \begin{subfigure}[b]{0.47\textwidth}
        \centering
        \includegraphics[width=1.0\textwidth]{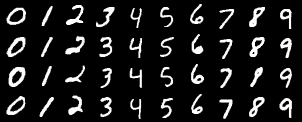}
        \caption{
            Conditional generation on MNIST.
        }
    \end{subfigure}
    \hfill
    \begin{subfigure}[b]{0.47\textwidth}
        \centering
        \includegraphics[width=1.0\textwidth]{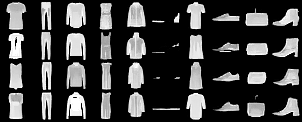}
        \caption{
            Conditional generation on Fashion-MNIST.
        }
    \end{subfigure}
    \caption{
        Testing conditional generation using MNIST-style datasets. Coherent
        samples demonstrate that the proposed conditioning method does inject class
        information.
    }
    \label{fig:mnist}
\end{figure}

Another critical component of an ideal generative model is the ability to
control its generation. We explore class-conditioned image generation of
ImageNet at $256 \times 256$ resolution, using pretrained ImageNet256
\gls{vqgan} checkpoints. 

There are many valid ways of injecting a conditioning signal into generative
models, for example passing one-hot or embedding class vectors. We use a simple
solution proposed in~\citet{parmar2018image}: adding a learned class embedding to
every input embedding. To test whether their proposed method can also be applied
to \gls{sundae}, we conducted an experiment on discrete MNIST-style datasets. We
treat each of the possible 8-bit greyscale colour values as a codebook
index, resulting in $\vqganNbLatents = 256$---generating pixels directly rather
than image patches. Results of these experiments are shown in
Figure~\ref{fig:mnist} and demonstrate that \gls{sundae} can incorporate
conditional information using this simple approach.

Despite this, our model fails to produce reasonable samples for all classes in
ImageNet. On classes containing large scenes such as landscapes, the samples
are convincing and diverse. However, for classes requiring fine-grained detail,
the outputs merely resemble the target class. Results of conditional generation
with four representative classes are shown in Figure~\ref{fig:imagenet}. Due to
this, we chose not to compute perceptual metrics for conditional experiments as
the sample quality was clearly insufficient via inspection alone. This could be
a result of lack of model capacity, lack of training time, or the conditioning
strategy tested on MNIST being insufficient for ImageNet. The training of a more
effective conditional model is left for future work.

\subsection{Arbitrary Image Inpainting}
\label{subsec:evaluationInpainting}

\begin{figure}[h]
    \centering
    \begin{subfigure}[b]{0.47\textwidth}
        \centering
        \includegraphics[width=1.0\textwidth]{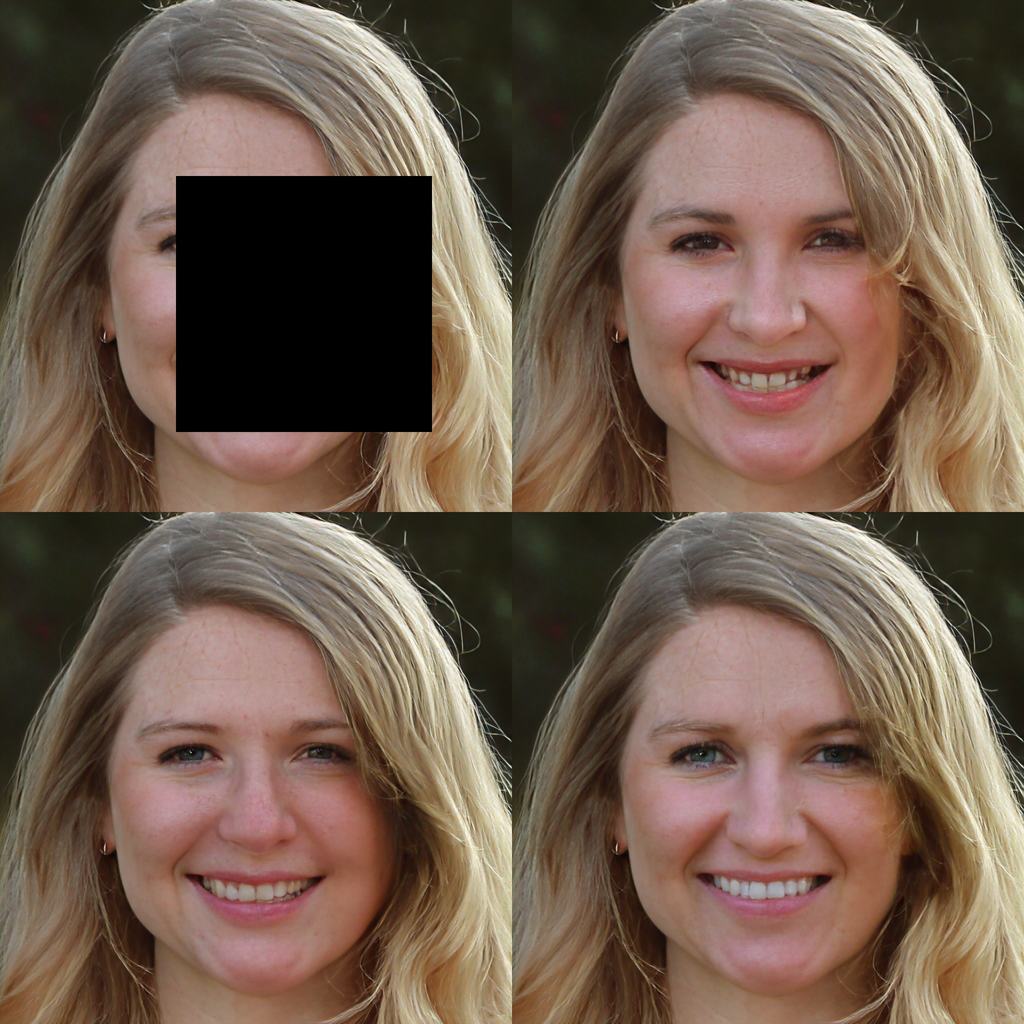}
        \caption{
            Multiple results of inpainting using the same block mask.
        }
    \end{subfigure}
    \hfill
    \begin{subfigure}[b]{0.47\textwidth}
        \centering
        \includegraphics[width=1.0\textwidth]{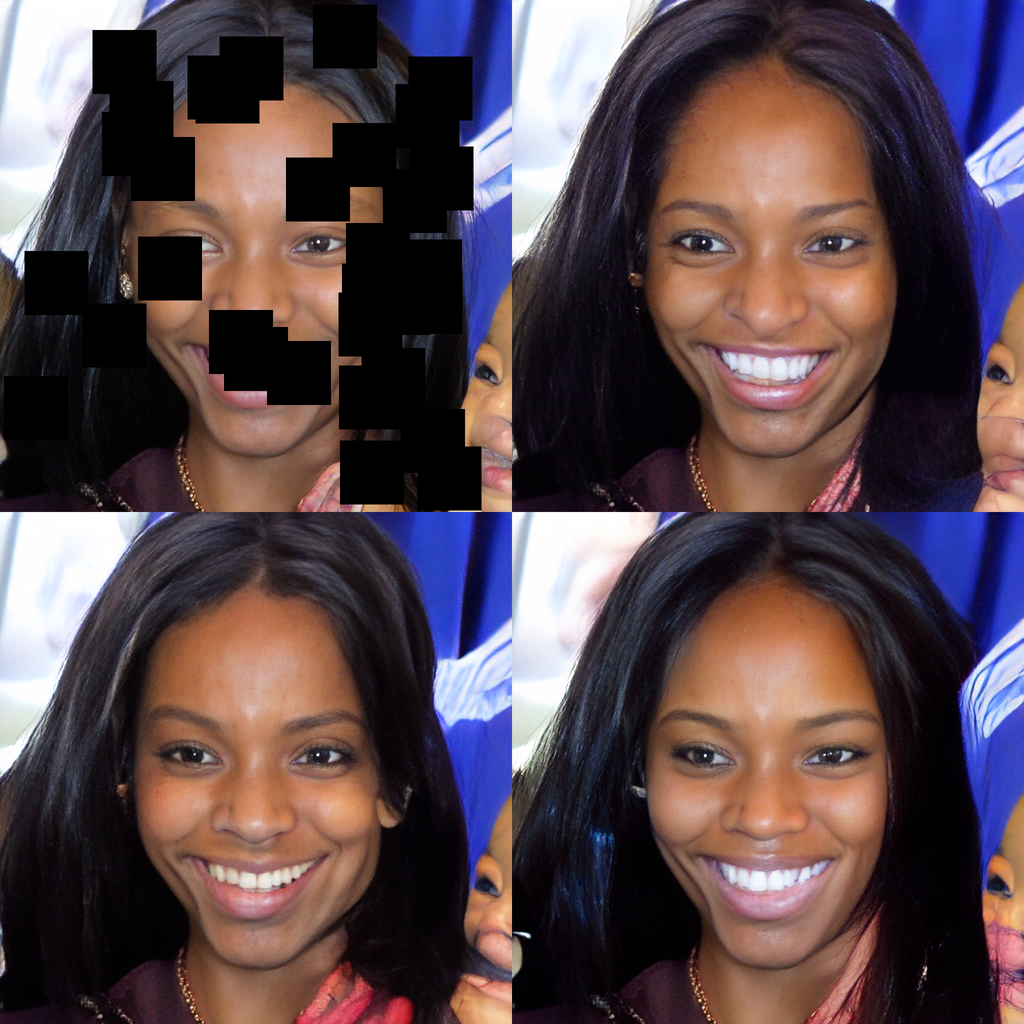}
        \caption{
            Multiple results of inpainting on the same random mask. 
        }
    \end{subfigure}
    \caption{
        Inpainting results on FFHQ1024. We compute multiple outputs per input
        image and mask to demonstrate diversity of outputs. 
    }
    \label{fig:inpaint}
\end{figure}

As outlined earlier, \acrlong{nar} generative models have a number of advantages
for inpainting tasks, including supporting arbitrary masks and being able to use
the full context available to them. We provide a number of examples of
inpainting on FFHQ1024, showcasing different patterns and results given the same
starting image and mask. As our method utilises a \gls{vq} image model, it is
incapable of doing fine-grained inpainting at a pixel level. We found in
practise this had little effect on the perceptual quality of the outputs, as
shown in Figure~\ref{fig:inpaint}.

\subsection{Limitations}
\label{subsec:evaluationLimitation}

As a result of our evaluation, some limitations of our approach arise. One
weakness is that our \gls{vq} image model utilises adversarial components within
it. This potentially means that image patches (corresponding to each codebook
entry) could suffer from mode collapse issues, which does indeed occur with
certain hair textures. Additionally, we encountered consistent instability
during training of the large \glspl{vqgan} which resulted in many failed
training runs. 
Further research into high compression \gls{vq} models that do not use
adversarial components remains an open and challenging area of research. When
such a \gls{vq} model is designed, it can easily be substituted into our
proposed framework.

Despite our model demonstrating extremely fast sampling it falls short of many
recent methods in terms of perceptual quality metrics. Though measures of
perceptual quality such as FID are known to be
flawed~\citep{chong2020effectively}, other measures such as density and coverage
also show inferior sample quality~\citep{ferjad2020icml}. This is especially
true on ImageNet where many classes merely resemble the target class---likely
due to lack of model capacity and training time. Despite this, the resulting
samples on the FFHQ and CelebA datasets are still very diverse and of good
perceptual quality. Further work, particularly extensive hyperparameter sweeps,
is needed to improve quality in terms of these perceptual metrics.

\section{Conclusion}
In this work we investigated pushing the efficiency of generative models using a
modified hourglass transformer and a \acrlong{nar} sampling method, following
the trend in generative modelling research of simultaneously improving quality
and speed of sampling using non-adversarial approaches. We found that the
combination of these techniques formed a fast image generation framework. To our
surprise, the proposed method was faster and more scaleable than expected, able
to be applied with ease to megapixel images, and generate samples at such
resolutions in seconds---considerably faster than existing non-adversarial
methods. Additionally, we found that the performance of hourglass transformers
on multidimensional data can be improved through adjustments to resampling and
positional embedding mechanisms---applicable in any task applying hierarchical
transformers to multidimensional data. We also demonstrated the scalability of
\gls{sundae} by applying it to sequences of length 1024---longer than evaluated
on in the original work. This demonstrates the superiority of the \acrlong{nar}
paradigm, and joins a rapidly growing space of research into their use as a
viable alternative to \acrlong{ar} frameworks. Additional research is needed
into better \gls{vq} image models and into a stronger conditional generative
model, as well as closing the gap between our proposed solution and existing
models in terms of perceptual quality metrics.

\subsubsection*{Acknowledgments}

This work made use of the facilities of the N8 Centre of Excellence in
Computationally Intensive Research (N8 CIR) provided and funded by the N8
research partnership and EPSRC (Grant No. EP/T022167/1). The Centre is
co-ordinated by the Universities of Durham, Manchester and York.

In addition, this work has used Durham University’s NCC cluster. NCC has been
purchased through Durham University’s strategic investment funds, and is
installed and maintained by the Department of Computer Science.

\bibliography{references}
\bibliographystyle{iclr2022_conference}

\pagebreak

\appendix

\section{Implementation Details}
\subsection{SUNDAE Implementation}
\begin{table}[ht]
    \centering
    \makebox[\textwidth]{\begin{tabular}{|c||c c||c||c c||c||}
    \hline
    \textbf{Dataset} & \textbf{FFHQ256} & \textbf{FFHQ1024} & \textbf{CelebA}
                     & \textbf{MNIST} & \textbf{FashionMNIST} &
                     \textbf{ImageNet} \\
    \hline
    Dataset Size & $60,000$ & $60,000$ & $30,000$ & $60,000$ & $60,000$ & $1.28$M \\
    Codebook Size & $1024$ & $8192$ & $1024$ & $256$ & $256$ & $1024$ \\
    Latent Shape & $16 \times 16$ & $32 \times 32$ & $16 \times 16$ & $28 \times
                 28$ & $28 \times 28$ & $16 \times 16$ \\
    Unroll Steps & 3 & 3 & 3 & 2 & 2 & 3 \\
    \hline
    Depth & $3-16-3$ & $2-12-2$ & $2-12-2$ & $2-8-2$ & $2-8-2$ & $3-14-3$\\
    Dimension & 1024 & 1024 & 1024 & 1024 & 1024 & 1024 \\
    Shorten Factor & 4 & 4 & 4 & 4 & 4 & 4 \\
    Attention Heads & 8 & 8 & 8 & 8 & 8 & 12 \\
    Resample Type & Linear & Linear & Linear & Linear & Linear & Linear \\
    \hline
    Classes &---&---&---& 10 & 10 & 1000 \\
    Class Dimension &---&---&---& 1024 & 1024 & 1024 \\
    \hline
    \end{tabular}}
    \caption{
        Table of parameters for all training experiments. Depth is
        represented as three numbers corresponding to number of layers before
        downsampling, number of downsampled layers, and number of layers after
        upsampling. The dataset size is the size of the training split of the
        dataset. The latent shape of MNIST experiments is exactly equal to the
        shape of $\image$, as for these experiments we operate directly on a
        (discrete) pixels.
    }
    \label{tab:parameters}
\end{table}

All \gls{sundae} models are trained using the \textit{Adam}
optimizer~\citep{kingma2014adam} as realised in its \textit{AdamW}
implementation in PyTorch~\citep{paszke2019pytorch}. All models and training
scripts are implemented in PyTorch, trained on a single 80 GiB Nvidia A100.
Similarly, computation of perceptual scores such as FID was done on this same
device. Inference for all models can be run on a consumer-grade GPU, in our case
a Nvidia GTX 1080Ti.
All parameters used for training the \gls{sundae} models are shown
in Table~\ref{tab:parameters}.

\subsection{VQ-GAN Implementation}

\begin{table}[ht]
    \centering
    \makebox[\textwidth]{\begin{tabular}{|c|c|}
    \hline
    \textbf{Codeword Dimension} & 256 \\
    \textbf{Number of Codewords} & 8192 \\
    \textbf{Encoder Channels} & [128, 128, 256, 256, 512, 512] \\
    \textbf{Downsampling Rate $\vqganDownsample$} & 32 \\
    \textbf{Number of Residual Blocks} & 1 \\
    \textbf{Attention Resolution} & $\leq 64$ \\
    \textbf{Dropout} & 0.0 \\
    \hline
    \textbf{Discriminator Start} & 15000 \\
    \textbf{Number of Discriminator Layers} & 3 \\
    \textbf{Discriminator Weight} & 0.5 \\
    \textbf{Codebook Weight} & 1.0 \\
    \hline
    \textbf{Local Batch Size} & 1 \\
    \textbf{Global Batch Size} & 4 \\
    \hline
    \end{tabular}}
    \caption{Table of parameters for \gls{vqgan} training at $1024 \times 1024$
    resolution.}
    \label{tab:vqganParameters}
\end{table}

Where possible, we used pretrained \gls{vqgan} checkpoints provided by
\citet{esser2021taming}, who originally introduced \gls{vqgan}. For FFHQ1024
experiments, we trained our own \gls{vqgan} using training scripts provided by
\citet{esser2021taming} on four 32 GiB Nvidia V100 GPUs. The chosen
hyperparameters for \gls{vqgan} for this task is shown in
Table~\ref{tab:vqganParameters}.

\section{Additional Results}
\subsection{Effect of Sampling Hyperparameters}
\label{sec:additionalParameters}

This section shows a non-comprehensive overview of the effect of varying
sample-time hyperparameters on image quality.

\begin{figure}[htb!]
    \centering
    \begin{subfigure}[b]{0.32\textwidth}
        \centering
        \resizebox{\textwidth}{!}{
            \begin{tikzpicture}
\begin{axis}[
y label style={at={(axis description cs:-0.15,1.0)},rotate=-90,anchor=south},
title={},
xlabel={Sampling steps $T$},
ylabel={FID $\downarrow$},
xmin=0, xmax=500,
ymin=0, ymax=120,
xtick={0,50,100,150,200,250,300,350,400,450,500},
ytick={0,20,40,60,80,100,120},
legend pos=north east,
ymajorgrids=true,
grid style=dashed,
]\addplot[color=black, mark=square]
coordinates {(10.0, 20.014929483621366)(20.0, 21.81219325894136)(50.0, 27.547737014526252)(100.0, 31.09500698456076)(256.0, 38.10217399953847)(512.0, 43.68606647698149)};
\addlegendentry{FFHQ256}
\addplot[color=red, mark=*]
coordinates {(20.0, 26.87004056541223)(50.0, 27.141302141763884)(100.0, 28.447290731888177)(256.0, 32.4124245659133)};
\addlegendentry{FFHQ1024}
\addplot[color=blue, mark=diamond]
coordinates {(10.0, 23.629221897597233)(20.0, 21.759445128115722)(50.0, 27.835341472216665)(100.0, 34.92139271764068)(256.0, 47.65245888683146)(512.0, 64.48901983178908)};
\addlegendentry{CelebA}
\end{axis}
\end{tikzpicture}
        }
        \caption{Steps vs. FID.}
    \end{subfigure}
    \begin{subfigure}[b]{0.32\textwidth}
        \centering
        \resizebox{\textwidth}{!}{
            \begin{tikzpicture}
\begin{axis}[
y label style={at={(axis description cs:-0.15,1.0)},rotate=-90,anchor=south},
title={},
xlabel={Sampling steps $T$},
ylabel={Coverage $\uparrow$},
xmin=0, xmax=500,
ymin=0.0, ymax=1.8,
xtick={0,50,100,150,200,250,300,350,400,450,500},
ytick={0.0,0.2,0.4,0.6000000000000001,0.8,1.0,1.2000000000000002,1.4000000000000001,1.6,1.8},
legend pos=north east,
ymajorgrids=true,
grid style=dashed,
]\addplot[color=black, mark=square]
coordinates {(10.0, 0.6573)(20.0, 0.6934)(50.0, 0.6448)(100.0, 0.5869)(256.0, 0.5)(512.0, 0.4238)};
\addlegendentry{FFHQ256}
\addplot[color=red, mark=*]
coordinates {(20.0, 0.505)(50.0, 0.4561)(100.0, 0.4066)(256.0, 0.3628)};
\addlegendentry{FFHQ1024}
\addplot[color=blue, mark=diamond]
coordinates {(10.0, 0.3234)(20.0, 0.3811)(50.0, 0.3395)(100.0, 0.3079)(256.0, 0.2375)(512.0, 0.1952)};
\addlegendentry{CelebA}
\end{axis}
\end{tikzpicture}
        }
        \caption{Steps vs. Coverage.}
    \end{subfigure}
    \begin{subfigure}[b]{0.32\textwidth}
        \centering
        \resizebox{\textwidth}{!}{
            \begin{tikzpicture}
\begin{axis}[
y label style={at={(axis description cs:-0.15,1.0)},rotate=-90,anchor=south},
title={},
xlabel={Sampling steps $T$},
ylabel={Density $\uparrow$},
xmin=0, xmax=500,
ymin=0.0, ymax=1.8,
xtick={0,50,100,150,200,250,300,350,400,450,500},
ytick={0.0,0.2,0.4,0.6000000000000001,0.8,1.0,1.2000000000000002,1.4000000000000001,1.6,1.8},
legend pos=north east,
ymajorgrids=true,
grid style=dashed,
]\addplot[color=black, mark=square]
coordinates {(10.0, 1.2086666666666666)(20.0, 1.3613)(50.0, 1.1937333333333333)(100.0, 1.0614)(256.0, 0.8514999999999999)(512.0, 0.6940333333333333)};
\addlegendentry{FFHQ256}
\addplot[color=red, mark=*]
coordinates {(20.0, 0.8288)(50.0, 0.6492333333333333)(100.0, 0.5502)(256.0, 0.45646666666666663)};
\addlegendentry{FFHQ1024}
\addplot[color=blue, mark=diamond]
coordinates {(10.0, 0.43720000000000003)(20.0, 0.5754999999999999)(50.0, 0.574)(100.0, 0.4669333333333333)(256.0, 0.41146666666666665)(512.0, 0.30526666666666663)};
\addlegendentry{CelebA}
\end{axis}
\end{tikzpicture}
        }
        \caption{Steps vs. Density.}
    \end{subfigure}
    \caption{
        Plots showing sample quality in terms of different metrics as
        number of sampling steps $\markovSteps$ increases. Counter-intuitively, the
        sample quality decreases with number of sampling steps, seen on
        all metrics and datasets. This indicates that naively scaling the number
        of sampling steps (without adjusting number of parameters) does not
        necessarily improve results.
    }
    \label{fig:step}
\end{figure}
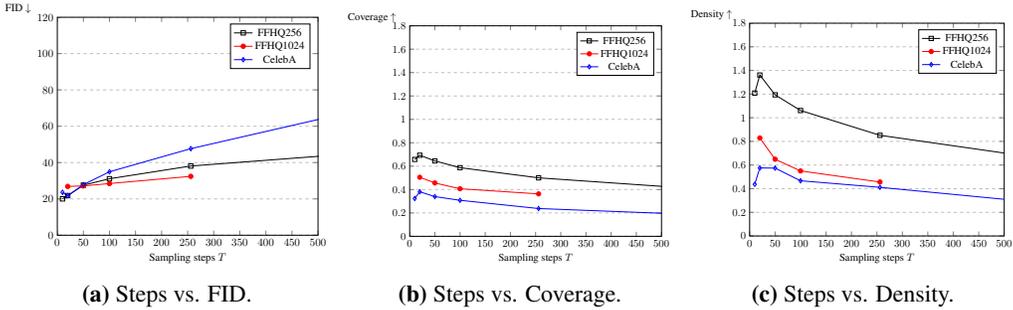

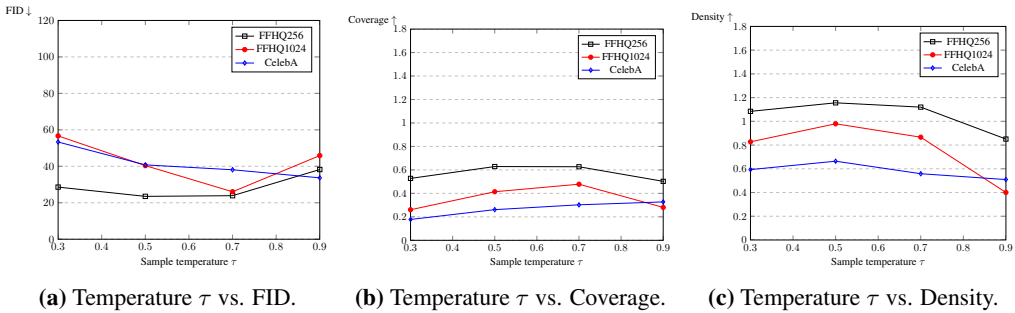
\begin{figure}[htb!]
    \begin{subfigure}[b]{0.32\textwidth}
        \centering
        \resizebox{\textwidth}{!}{
            \begin{tikzpicture}
\begin{axis}[
y label style={at={(axis description cs:-0.15,1.0)},rotate=-90,anchor=south},
title={},
xlabel={Sample temperature $\tau$},
ylabel={FID $\downarrow$},
xmin=0.3, xmax=0.9000000000000001,
ymin=0, ymax=120,
xtick={0.3,0.4,0.5,0.6000000000000001,0.7000000000000002,0.8000000000000003,0.9000000000000001},
ytick={0,20,40,60,80,100,120},
legend pos=north east,
ymajorgrids=true,
grid style=dashed,
]\addplot[color=black, mark=square]
coordinates {(1.0, 64.37565210300971)(0.9, 38.23760131680953)(0.7, 23.88038701179043)(0.5, 23.512077014725186)(0.3, 28.584934381185647)};
\addlegendentry{FFHQ256}
\addplot[color=red, mark=*]
coordinates {(1.0, 126.90269436417385)(0.9, 45.87416713841175)(0.7, 26.031535023121656)(0.5, 40.37608534196749)(0.3, 56.68016068363152)};
\addlegendentry{FFHQ1024}
\addplot[color=blue, mark=diamond]
coordinates {(1.0, 46.90379708867538)(0.9, 33.66781555558217)(0.7, 38.10430631800378)(0.5, 40.77642117557276)(0.3, 53.326141611079166)};
\addlegendentry{CelebA}
\end{axis}
\end{tikzpicture}
        }
        \caption{Temperature $\temperature$ vs. FID.}
    \end{subfigure}
    \begin{subfigure}[b]{0.32\textwidth}
        \centering
        \resizebox{\textwidth}{!}{
            \begin{tikzpicture}
\begin{axis}[
y label style={at={(axis description cs:-0.15,1.0)},rotate=-90,anchor=south},
title={},
xlabel={Sample temperature $\tau$},
ylabel={Coverage $\uparrow$},
xmin=0.3, xmax=0.9000000000000001,
ymin=0.0, ymax=1.8,
xtick={0.3,0.4,0.5,0.6000000000000001,0.7000000000000002,0.8000000000000003,0.9000000000000001},
ytick={0.0,0.2,0.4,0.6000000000000001,0.8,1.0,1.2000000000000002,1.4000000000000001,1.6,1.8},
legend pos=north east,
ymajorgrids=true,
grid style=dashed,
]\addplot[color=black, mark=square]
coordinates {(1.0, 0.2885)(0.9, 0.5026)(0.7, 0.6269)(0.5, 0.6286)(0.3, 0.5276)};
\addlegendentry{FFHQ256}
\addplot[color=red, mark=*]
coordinates {(1.0, 0.0865)(0.9, 0.2805)(0.7, 0.4781)(0.5, 0.4139)(0.3, 0.2606)};
\addlegendentry{FFHQ1024}
\addplot[color=blue, mark=diamond]
coordinates {(1.0, 0.1947)(0.9, 0.3274)(0.7, 0.3025)(0.5, 0.2614)(0.3, 0.1777)};
\addlegendentry{CelebA}
\end{axis}
\end{tikzpicture}
        }
        \caption{Temperature $\temperature$ vs. Coverage.}
    \end{subfigure}
    \begin{subfigure}[b]{0.32\textwidth}
        \centering
        \resizebox{\textwidth}{!}{
            \begin{tikzpicture}
\begin{axis}[
y label style={at={(axis description cs:-0.15,1.0)},rotate=-90,anchor=south},
title={},
xlabel={Sample temperature $\tau$},
ylabel={Density $\uparrow$},
xmin=0.3, xmax=0.9000000000000001,
ymin=0.0, ymax=1.8,
xtick={0.3,0.4,0.5,0.6000000000000001,0.7000000000000002,0.8000000000000003,0.9000000000000001},
ytick={0.0,0.2,0.4,0.6000000000000001,0.8,1.0,1.2000000000000002,1.4000000000000001,1.6,1.8},
legend pos=north east,
ymajorgrids=true,
grid style=dashed,
]\addplot[color=black, mark=square]
coordinates {(1.0, 0.40446666666666664)(0.9, 0.8507333333333333)(0.7, 1.1197333333333332)(0.5, 1.156)(0.3, 1.0841666666666665)};
\addlegendentry{FFHQ256}
\addplot[color=red, mark=*]
coordinates {(1.0, 0.1174)(0.9, 0.4005666666666666)(0.7, 0.8662666666666667)(0.5, 0.9793333333333334)(0.3, 0.8272666666666666)};
\addlegendentry{FFHQ1024}
\addplot[color=blue, mark=diamond]
coordinates {(1.0, 0.18)(0.9, 0.5093666666666666)(0.7, 0.5584)(0.5, 0.6640333333333333)(0.3, 0.5931666666666666)};
\addlegendentry{CelebA}
\end{axis}
\end{tikzpicture}
        }
        \caption{Temperature $\temperature$ vs. Density.}
    \end{subfigure}
    \caption{
        Plots showing sample quality in terms of different metrics as sample
        temperature $\temperature$ is changed. Given the other parameters, a good choice
        of $\temperature$ falls in the range $0.5-0.7$. However, this range may
        differ depending on the values of other parameters, and naturally higher
        temperatures lead to more diverse samples.
    }
    \label{fig:temp}
\end{figure}

\FloatBarrier

\pagebreak
\subsection{Unconditional Samples}
\label{sec::additionalFFHQ}

\begin{figure}[htb!]
    \centering
    \makebox[\textwidth]{\includegraphics[width=1.1\textwidth]{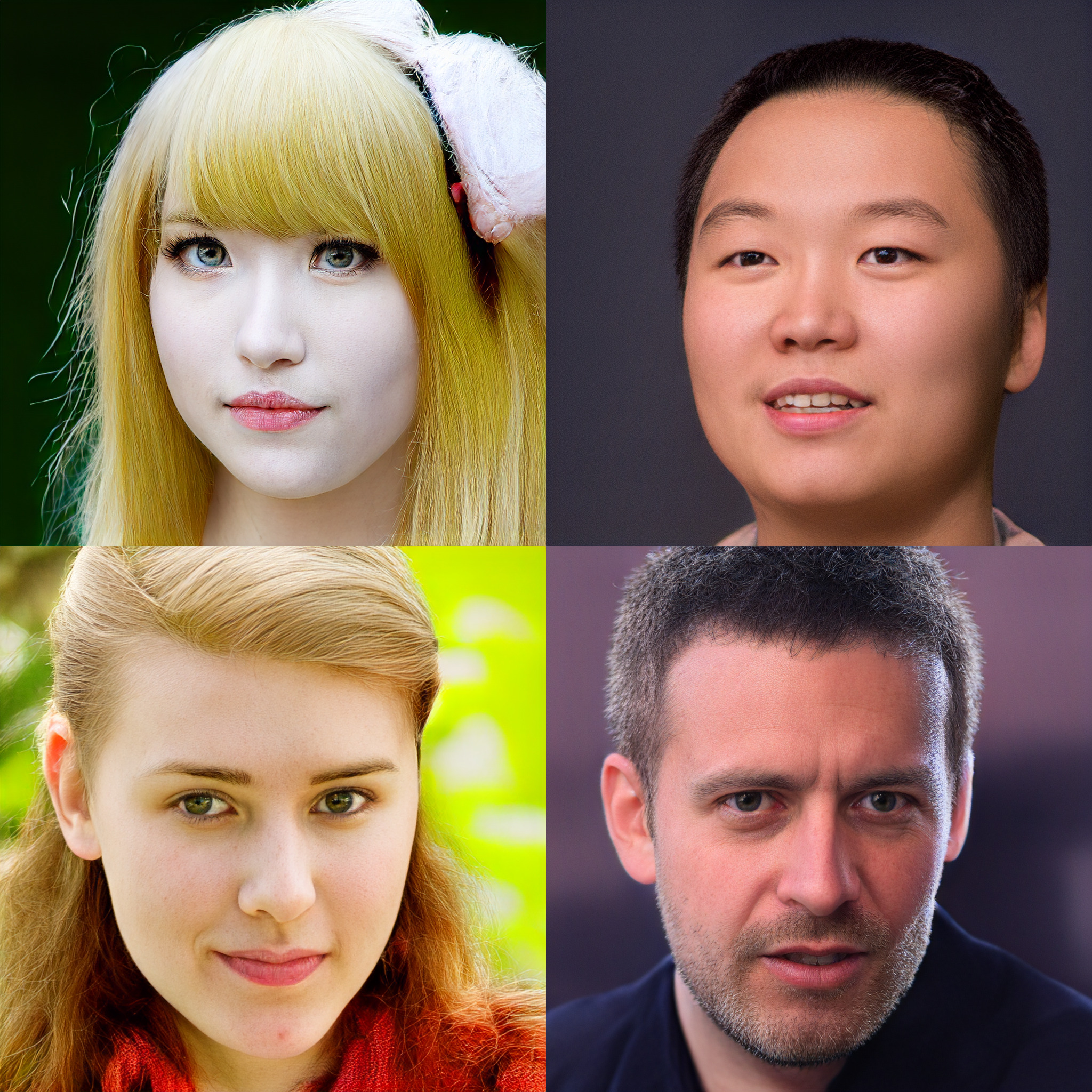}}
    \caption{
        Unconditional FFHQ $1024 \times 1024$ samples generated
        in $\markovSteps = 100$ sampling steps.
    }
\end{figure}

\begin{figure}[htb!]
    \centering
    \makebox[\textwidth]{\includegraphics[width=1.1\textwidth]{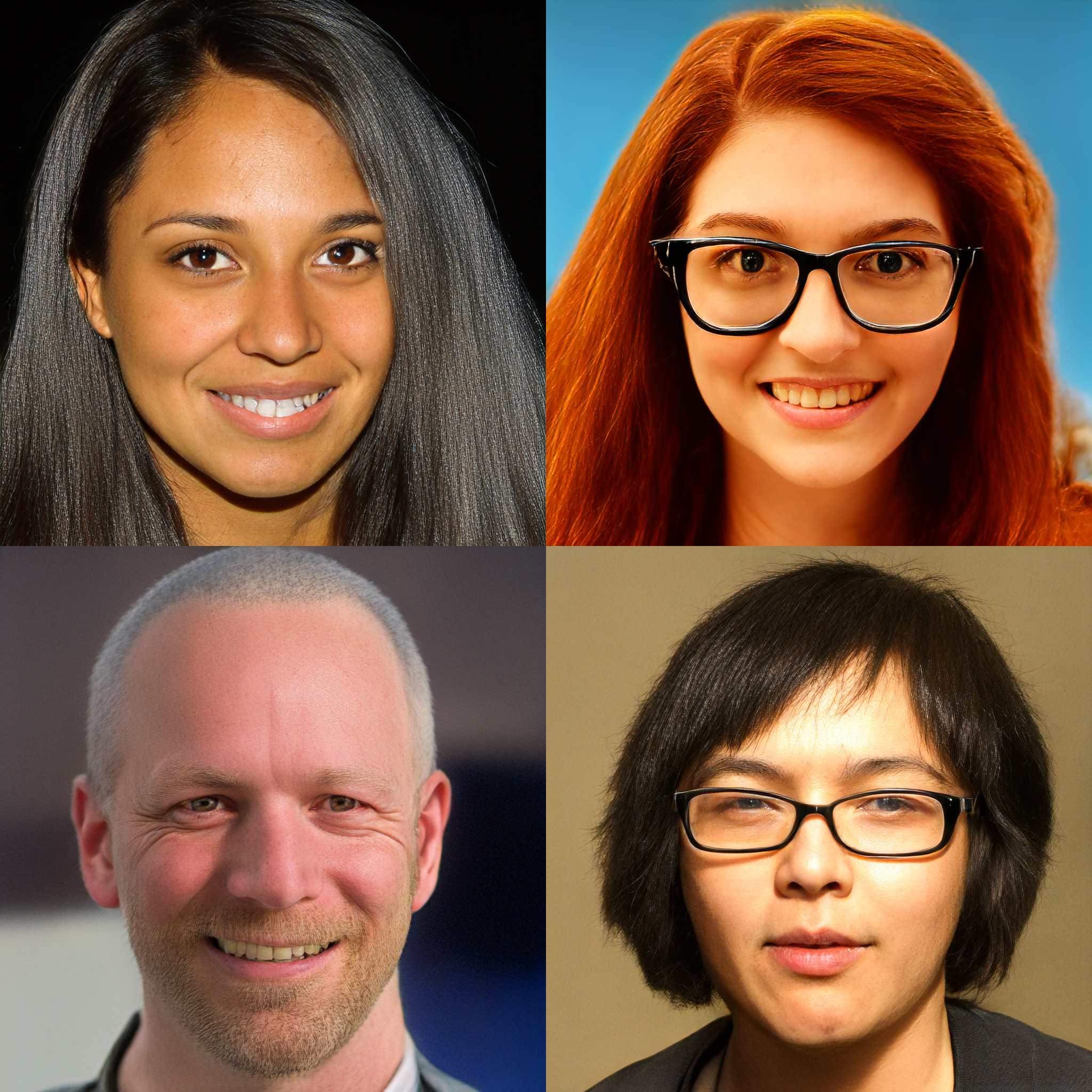}}
    \caption{
        Unconditional FFHQ $1024 \times 1024$ samples generated
        in $\markovSteps = 100$ sampling steps.
    }
\end{figure}

\begin{figure}[htb!]
    \centering
    \makebox[\textwidth]{\includegraphics[width=1.1\textwidth]{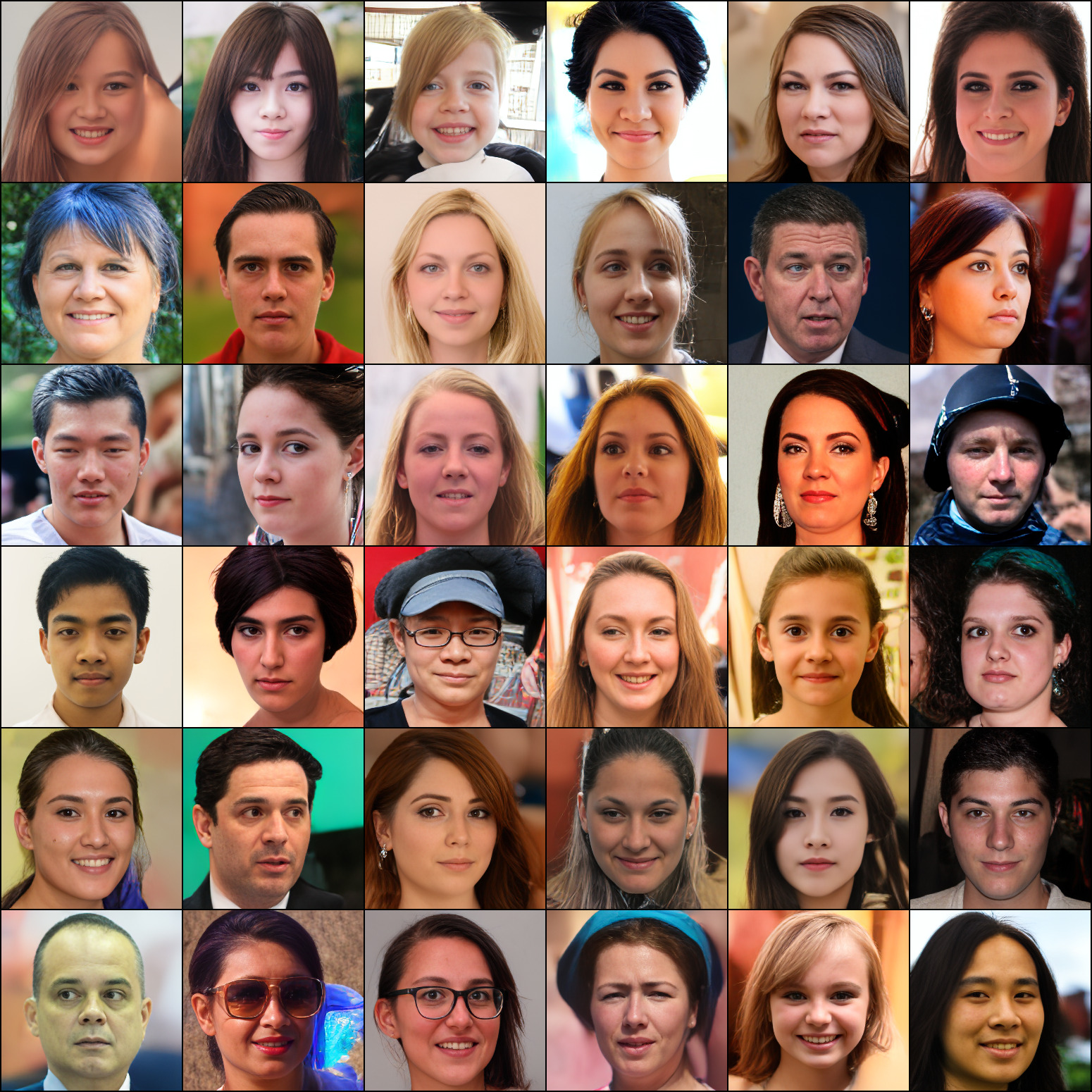}}
    \caption{
        Unconditional FFHQ $256 \times 256$ samples generated
        in $\markovSteps = 50$ sampling steps.
    }
\end{figure}

\begin{figure}[htb!]
    \centering
    \makebox[\textwidth]{\includegraphics[width=1.1\textwidth]{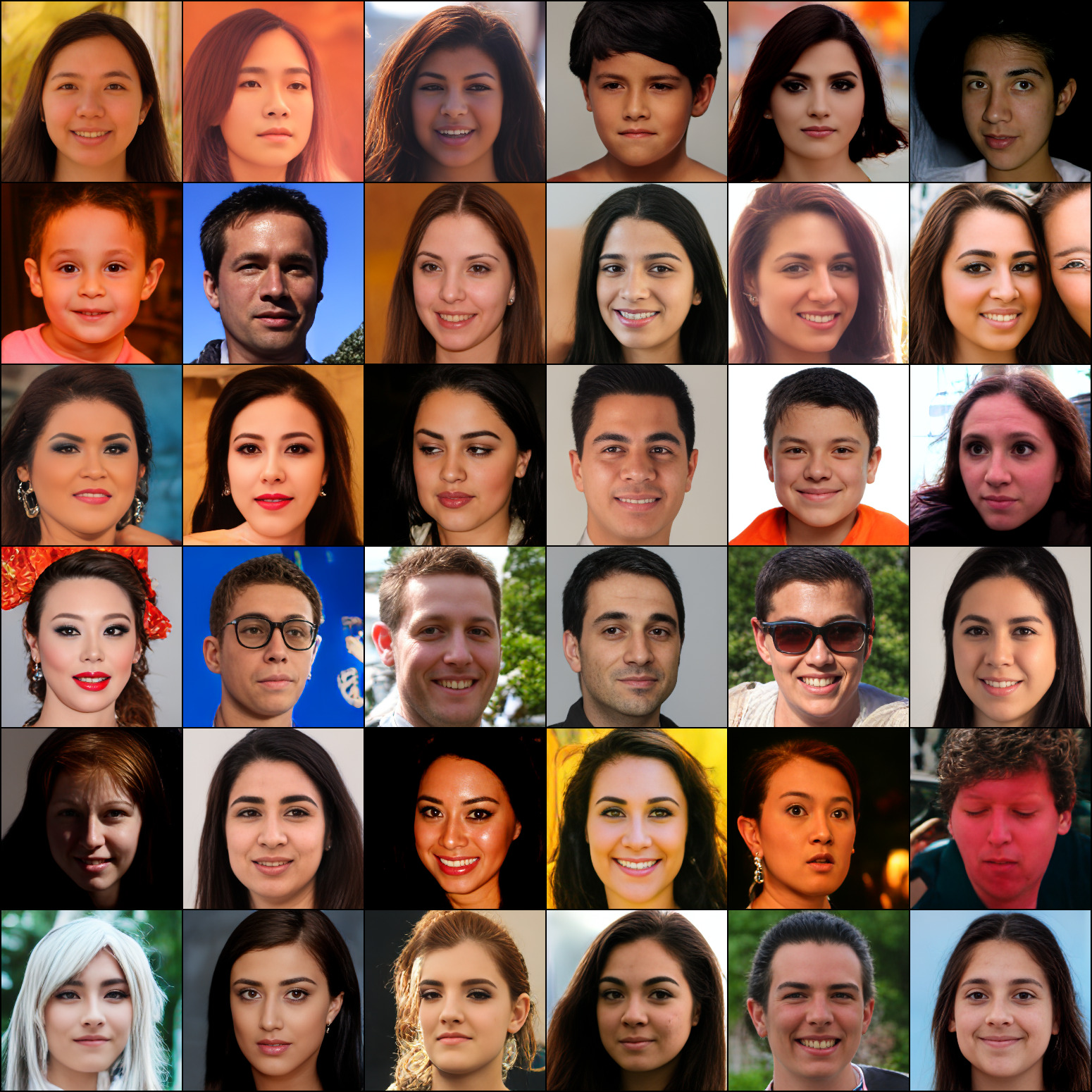}}
    \caption{
        Unconditional FFHQ $256 \times 256$ samples generated
        in $\markovSteps = 100$ sampling steps.
    }
\end{figure}

\begin{figure}[htb!]
    \centering
    \makebox[\textwidth]{\includegraphics[width=1.1\textwidth]{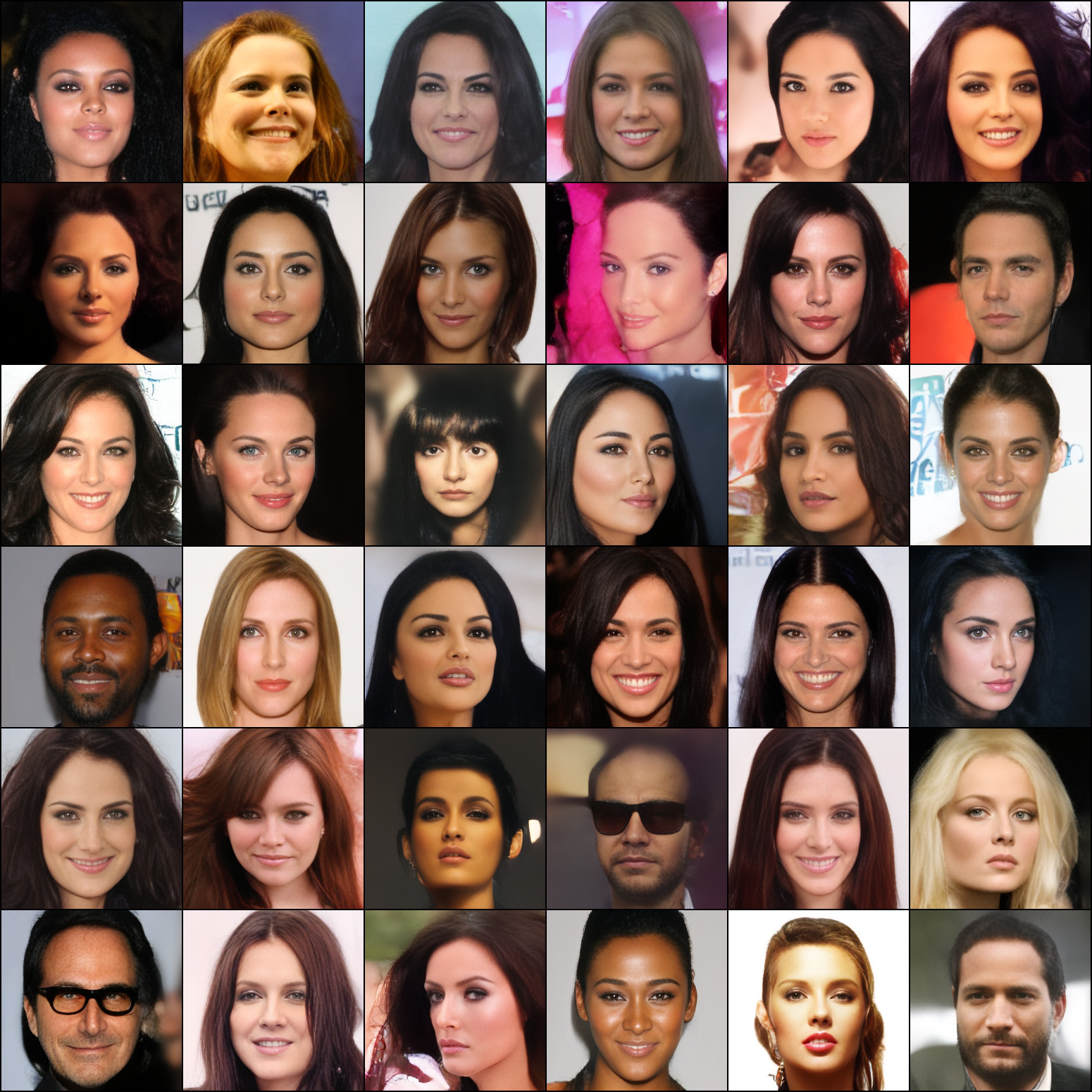}}
    \caption{
        Unconditional CelebA $256 \times 256$ samples
        generated in $\markovSteps = 50$ sampling steps.
    }
\end{figure}

\begin{figure}[htb!]
    \centering
    \makebox[\textwidth]{\includegraphics[width=1.1\textwidth]{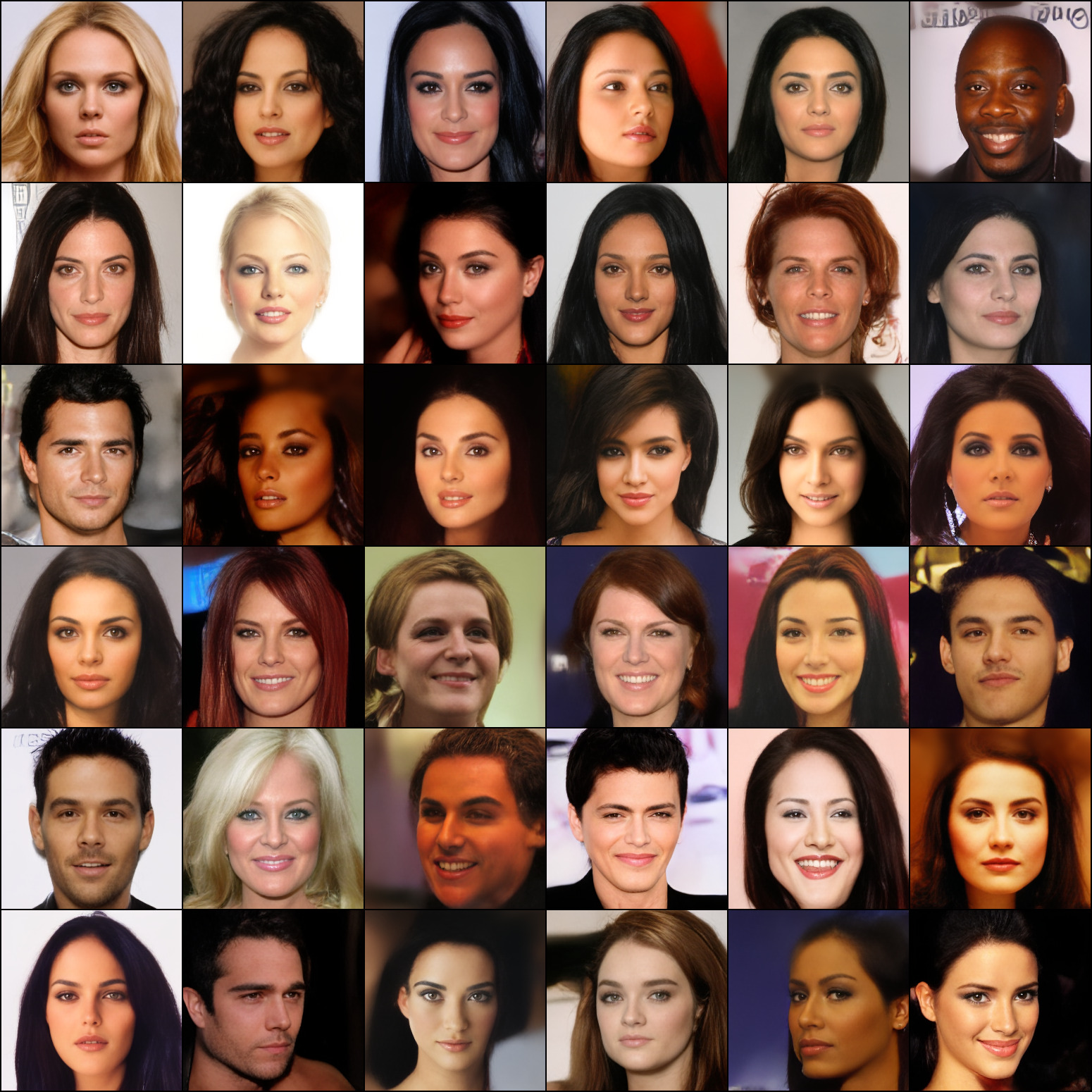}}
    \caption{
        Unconditional CelebA $256 \times 256$ samples
        generated in $\markovSteps = 100$ sampling steps.
    }
\end{figure}

\FloatBarrier
\subsection{Image Inpainting}

\begin{figure}[htb!]
    \centering
    \makebox[\textwidth]{\includegraphics[width=1.1\textwidth]{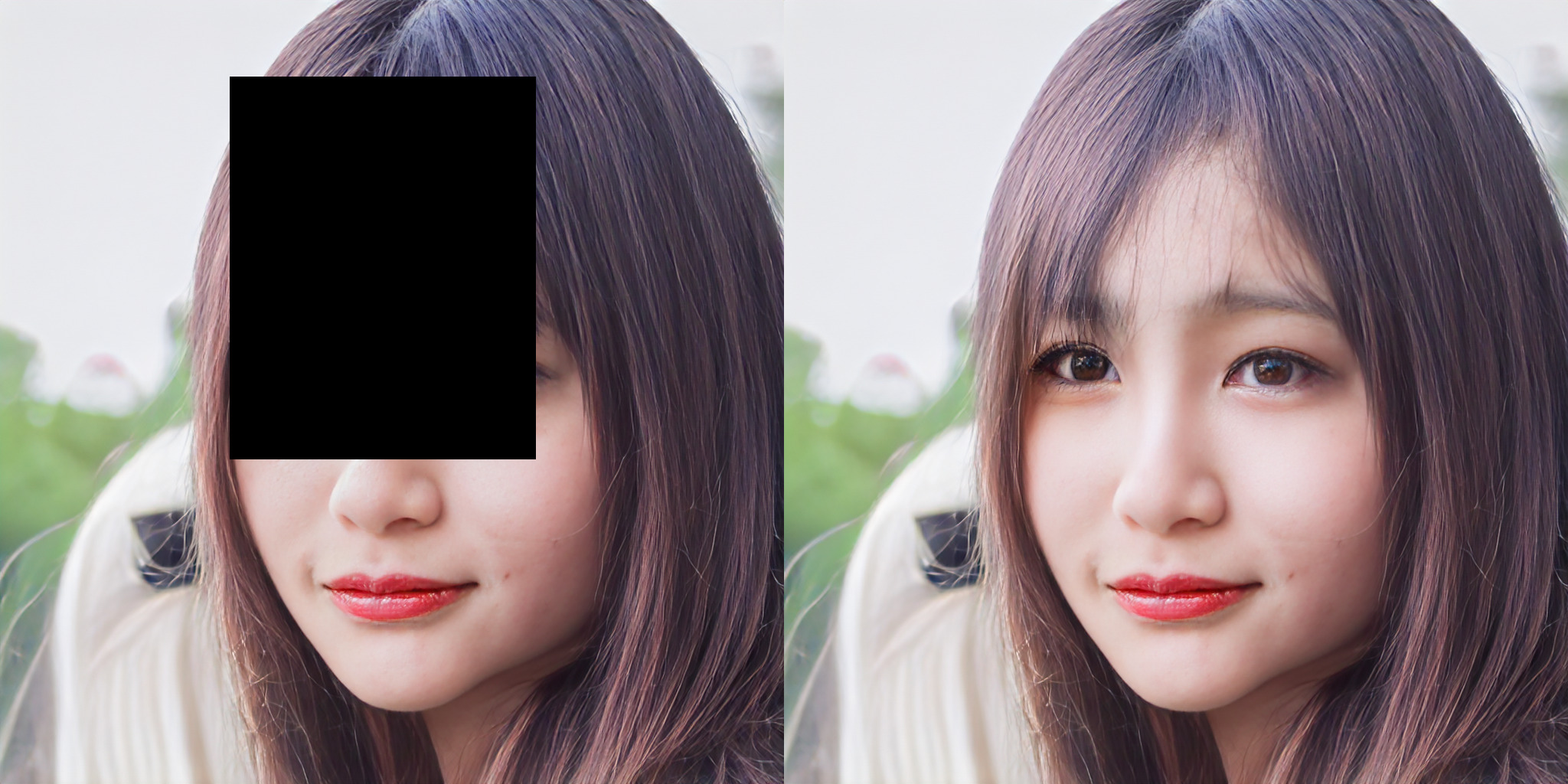}}
    \caption{
        Large example of inpainting at $1024 \times 1024$ resolution using FFHQ
        model.
    }
\end{figure}

\end{document}